\documentclass[lettersize,journal]{IEEEtran}
\usepackage{amsmath,amsfonts}
\usepackage{algorithmic}
\usepackage{algorithm}
\usepackage{array}
\usepackage[caption=false,font=normalsize,labelfont=sf,textfont=sf]{subfig}
\usepackage{textcomp}
\usepackage{stfloats}
\usepackage{url}
\usepackage{verbatim}
\usepackage{graphicx}
\usepackage{cite}
\usepackage{color}
\usepackage{multirow}
\usepackage{times}
\usepackage{epsfig}
\usepackage{amsmath}
\usepackage{amssymb}
\usepackage{float}
\usepackage{siunitx}
\usepackage{booktabs}
\usepackage{color}
\usepackage{array}
\usepackage{balance}
\usepackage{makecell}
\begin{document}

\title{Multiple Latent Space Mapping for Compressed Dark Image Enhancement}

\author{Yi Zeng,
        Zhengning Wang$^*$,
        Yuxuan Liu,
        Tianjiao Zeng,
        Xuhang Liu,\\
        Xinglong Luo,
        Shuaicheng Liu,
        Shuyuan Zhu,
        Bing Zeng~\IEEEmembership{Fellow,~IEEE}

\thanks{Yi Zeng, Zhengning Wang Yuxuan Liu, Tianjiao Zeng, Xuhang Liu, Xinglong Luo, Shuaicheng Liu, Shuyuan Zhu, Bing Zeng are with the School of Information and Communication Engineering, University of Electronic Science and Technology of China,
Cheng Du, China, 611731 e-mail: zhengning.wang@uestc.edu.cn}
}



\maketitle

\begin{abstract}
Dark image enhancement aims at converting dark images to normal-light images. Existing dark image enhancement methods take uncompressed dark images as inputs and achieve great performance. However, in practice, dark images are often compressed before storage or transmission over the Internet. Current methods get poor performance when processing compressed dark images. Artifacts hidden in the dark regions are amplified by current methods, which results in uncomfortable visual effects for observers. Based on this observation, this study aims at enhancing compressed dark images while avoiding compression artifacts amplification. Since texture details intertwine with compression artifacts in compressed dark images, detail enhancement and blocking artifacts suppression contradict each other in image space. Therefore, we handle the task in latent space. To this end, we propose a novel latent mapping network based on variational auto-encoder (VAE). Firstly, different from previous VAE-based methods with single-resolution features only, we exploit multiple latent spaces with multi-resolution features, to reduce the detail blur and improve image fidelity. Specifically, we train two multi-level VAEs to project compressed dark images and normal-light images into their latent spaces respectively. Secondly, we 
leverage a latent mapping network to transform features from compressed dark space to normal-light space. 
Specifically, since the degradation models of darkness and compression are different from each other, the latent mapping process is divided mapping into enlightening branch and deblocking branch. Comprehensive experiments demonstrate that the proposed method achieves state-of-the-art performance in compressed dark image enhancement.
\end{abstract}


\section{Introduction}
\label{intro} 
\IEEEPARstart{D}{ark} image enhancement is a basic task in image processing that aims to convert a dark image to a normal-light image for better visual effects. Up till now, many efforts have been made for dark image enhancement. Retinex-based methods like \cite{guo2016lime, zhuang2021bayesian} estimate the illumination map to attain reflectance map, and regard reflectance map as the enhanced result. In recent years, numerous neural networks concerning illumination enhancement have been proposed. Deep learning methods \cite{Chen2018Retinex, zhang2019kindling} enhance images by decomposing dark images into reflectance map and illumination map based on Retinex theory. Jiang \textit{et. al.} \cite{jiang2021enlightengan} adopts generative adversarial networks to make under-exposed images enlightening. These methods take uncompressed images as inputs and achieve great performance. However, in actual applications, owing to the enormous data of the raw images, compression and encoding operation would be taken on images before transmission at the cost of information loss. Processing compressed images with existing dark image enhancement algorithms may amplify the compression blocking artifacts and causes poor visual effects, which is analyzed as following.


\begin{figure}
    \centering
    \includegraphics[scale=0.82]{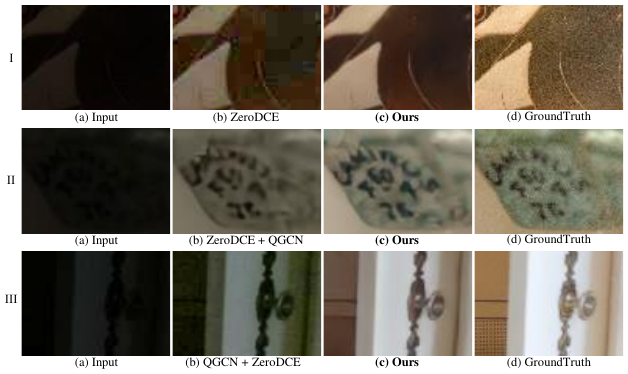}
    \caption{\textbf{Zoom-in details of the enhanced compressed dark image.} (I-a, II-a, III-a) show the original images. (I-b) shows the comparative result with \textbf{enhancement} method. (II-b) shows the comparative result with the combination of the enhancement method and deblocking method \textbf{(enhancement+deblocking)}. (III-b) shows the comparative result with the combination of deblocking method and enhancement method \textbf{(deblocking+enhancement)}. ZeroDCE\cite{guo2020zero} is a enhancement method and QGCN\cite{li-2020} is a deblocking method.}
    \label{fig:1}
\end{figure}

In the compression coding system such as JPEG which is based on discrete cosine transformation (DCT), the input image is partitioned into $8\times8$ encoding blocks and then DCT is implemented in each encoding block individually. The DCT coefficients will be quantized to reduce storage space. Such information loss weakens the correlation between different encoding blocks, causing discontinuities at the block boundaries. 
In the dark region, as texture details coupled with compression artifacts, existing dark image enhancement methods amplify the compression blocking artifacts hidden in the dark region, and amplify color distortion (adjacent encoding blocks expose the large difference in color), which is shown in Fig.\ref{fig:1} (I.b). Although numerous approaches for compression artifacts suppression \cite{dong2015compression, fu2021model, lin2019deep, li-2020} have been developed, the discontinuity between neighbor blocks will be amplified dramatically and hard to smooth via post-processing. Fig.\ref{fig:1} (II.b) shows the result of \textbf{enhancement+deblocking}, \textit{i.e.}, performing the enhancement algorithm, and then the deblocking algorithm afterwards. Obviously, texture detail in Fig.\ref{fig:1} (II.b) suffer from distortion.
In addition, since the blocking artifacts are hard to perceive in dark compressed images, direct processing deblocking algorithms could result in blurred details. Fig.\ref{fig:1} (III.b) shows the result of \textbf{deblocking+enhancement}, \textit{i.e.}, performing the deblocking algorithm, and then the enhancement algorithm. It can be seen some blocking artifacts in Fig.\ref{fig:1} (III.b) are still remained.
\begin{figure}
    \centering
    \includegraphics[scale=0.52]{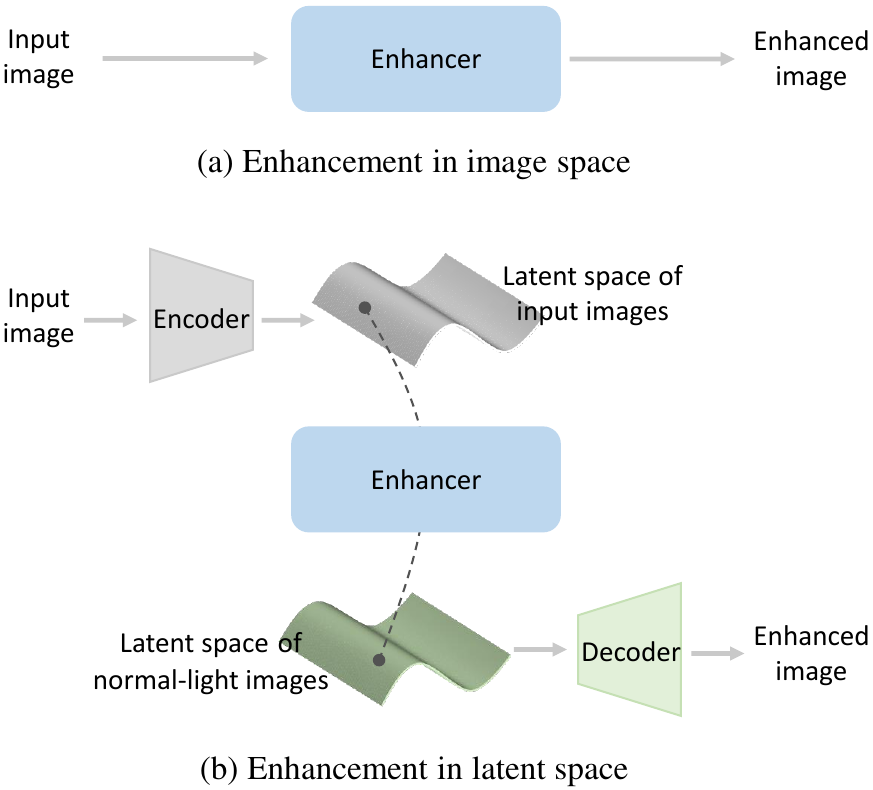}
    \caption{\textbf{Conceptual comparison of two enhancement mechanisms.} Image space enhancement enhances from image to image directly, and latent space enhancement work through mapping points on manifolds.}
    \label{fig:hidden_space_mapping}
\end{figure}

\begin{figure}
    \centering
    \includegraphics[scale=0.17]{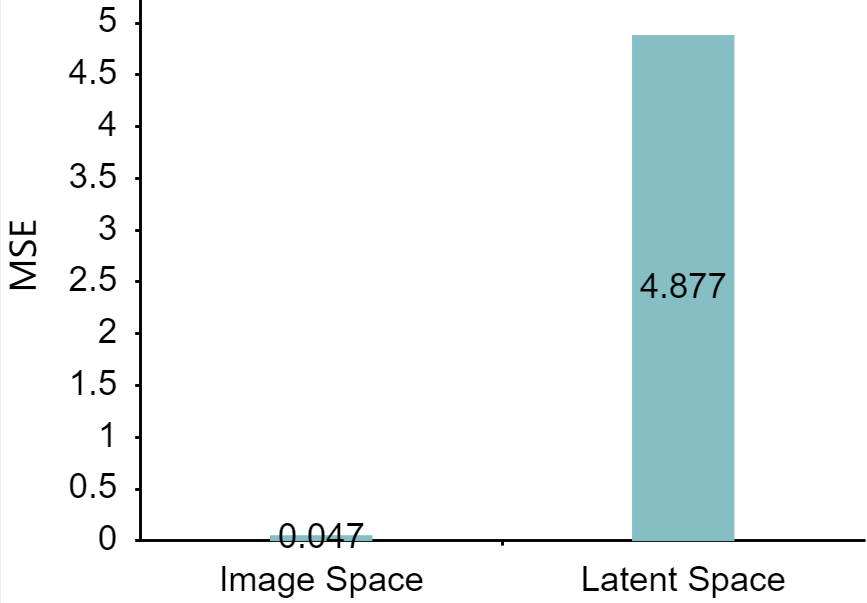}
    \caption{\textbf{Mean Square Error (MSE) between compressed data and uncompressed data on dark face test dataset with 6000 dark images.} 
    }
    \label{fig:MSE}
\end{figure}

Most methods perform enhancement in image space, enhancing the dark images to enhanced results directly (shown in Fig.\ref{fig:hidden_space_mapping} (a)). However, in compressed dark images, the blocking artifact is a type of detail. Since the texture detail intertwines with blocking artifacts, detail enhancement and blocking artifacts suppression contradict each other in image space. On the other hand, as data representation in latent space is denser than in image space \cite{gopalakrishnan2022classify}, learning a mapping model in latent space is easier than in image space \cite{wan2020bringing}. 
Besides, the difference between compressed dark and uncompressed dark images is too small, so that it is difficult for enhancer to discriminate the blocking artifacts hidden in compressed dark images. We argue that better discrimination between the compressed and uncompressed data is presented in latent space than in image space, and we found that Mean Square Error (MSE) between compressed data and uncompressed data is rise after projecting into latent space (0.047 v.s. 4.877) as illustrated in Fig.\ref{fig:MSE}. Therefore, our approach takes the mapping from compressed dark images to normal-light images in latent space as shown in Fig.\ref{fig:hidden_space_mapping} (b).


In this study, we propose a novel latent mapping network based on variational auto-encoder (VAE). Firstly, different from previous VAE-based methods which map with single-resolution features only, we propose to learn multiple latent spaces with multi-resolution features, to maintain detailed resolution information and improve semantic representation ability. Specifically, we train two multi-level VAEs to project compressed dark images and normal-light images into latent spaces respectively. Secondly, for each feature-level, we train a latent mapping network to transform features from compressed dark space to normal-light space, to learn the restoration from compressed dark images to normal-light images. For the latent space mapping, we divide it into two branches, \textit{i.e.,} enlighten branch and deblocking branch, as the degradation models of darkness and compression are different from each other. 
Comprehensive experiments demonstrate that the proposed method achieves state-of-the-art performance in compressed dark image enhancement.


Our contributions are summarized as follows.
\begin{itemize}
 \item We introduce the compressed dark image enhancement task that aims to enhance compressed dark images while avoiding amplifying the compression artifacts. We take this task as a mapping problem and propose mapping in latent space to reduce the difficulty of learning an enhancer.
 \item Our approach learns multiple latent spaces with multi-resolution features to reduce the detail blur and improve image fidelity.
 \item In this paper, we divide the latent space mapping into two branches, \textit{i.e.,} enlighten branch and deblocking branch, as the degradation models of darkness and compression are different from each other.
 \item Comprehensive experiments demonstrate that the proposed method achieves state-of-the-art performance in compressed dark image enhancement.
\end{itemize}

\section{Related Work}
\subsection{Conventional Methods}

Histogram equalization (HE) enhances the images by stretching the dynamic range. The variants of HE achieve some improvements, such as \cite{ibrahim2007brightness} can preserve the mean brightness of the input image. Another extension \cite{abdullah2007dynamic} stretches the dynamic range without making any loss of details in it. Wang \textit{et. al.} \cite{wang2007fast} propose a fast and effective method by modifying weight and threshold. In the conventional field, one of the most widely employed methods is based on Retinex theory, which decomposes an image into reflectance map and the illumination map, and uses reflectance map as the final enhanced result. \cite{jobson1997properties} estimates illumination by single-scale Gaussian blur. In \cite{zhang2018high}, an exposure correction based on Retinex is introduced. 
Fu \textit{et. al.} \cite{fu2016weighted} estimates both the reflectance and the illumination by a weighted variational model. Guo \textit{et. al.} \cite{guo2016lime} propose estimating illumination by imposing a structure prior to it. Methods \cite{li2018structure, ren2020lr3m} pay attention to noise suppression. Li \textit{et. al.} \cite{li2018structure} introduces a robust Retinex model that considers a noise map, to improve the enhancement performance of low-light images accompanied by intensive noise. And Ren \textit{et. al.} \cite{ren2020lr3m} is based on the low-rank regularized Retinex Model, which is called LR3M, to suppress noise in the reflectance map.

\subsection{Deep-learning Methods}
In recent years, more researchers have focused on deep learning. Inspired by bilateral grid processing, Gharbi \textit{et. al.} \cite{gharbi2017deep} introduces a neural network architecture. In \cite{ren2019low}, a hybrid network with an encoder-decoder network and RNN was applied to improve image quality. Hu \textit{et. al.} \cite{hu2018exposure} presents an automatic retouching system learning from unpaired data, by modeling retouching operations in a unified manner as resolution-independent differentiable filters. Methods \cite{wang2019underexposed, Chen2018Retinex, zhang2019kindling, liu2021retinex} construct their network based on Retinex theory. DRBN \cite{yang2020fidelity} recovers a linear band representation with paired images, and then recomposes the given bands via another learnable linear transformation with paired images, to extract a series of coarse-to-fine band representations with unpaired images. In \cite{ xu2020learning}, Xu \textit{et. al.} propose a frequency-based decomposition-enhancement model, which learns to recover image in the low-frequency layer and then enhances details in high-frequency. \cite{li2021low} repeatedly unfold the input image for feature extraction by a recursive unit that is composed of a recursive layer and a residual block. Xia \textit{et. al.} \cite{xia2021deep} introduce a neural network that applies the kernel field to the no-flash image, and then multiplies the result with the gain map to create the final output, to denoise low-light images. Different from the single image enhancement, \cite{cai2018learning} takes advantage of multi-exposure images to adjust brightness. Moseley \textit{et. al.}  \cite{moseley2021extreme} present a denoising approach to enhance extremely low-light images of permanently shadowed regions (PSRs) on the lunar surface.  Some researchers adopted generative adversarial networks for image enhancement processing, such as
\cite{chen2018deep, jiang2021enlightengan,  wang2019rdgan}.  Guo \textit{et. al.} \cite{guo2020zero} propose a self-supervised approach for image enhancement, and this work formulates light enhancement as a task of image-specific curve estimation.\cite{xu2022structure} propose a structure-texture aware network, which is called STANet,  to improve perceptual quality through successfully exploiting structure and texture features of low-light images. Ma \textit{et. al.} \cite{ma2022toward} develop
a self-calibrated illumination learning framework, to realize a faster, more flexible, and more robust enhancement model. These approaches attain remarkable achievements in image enhancement, but most of them rely on high-quality images. Side effects such as blocking artifacts amplification arise when dealing with compressed image through these algorithms.

\subsection{Artifacts Reduction}
Blocking artifacts bring terrible visual effect to observers. To reduce blocking artifacts and obtain high-quality images, researchers have proposed many solutions. Various compression standard, \textit{i.e.}, JPEG\cite{wallace1992jpeg}, JPEG2000\cite{skodras2001jpeg}, H.264\cite{wiegand2003overview}, \textit{etc.}, adopt deblocking filter in the decoder. The early work ARCNN \cite{dong2015compression} first adopts CNN to solve this problem, and create a precedent for the CNN-based methods. In \cite{zhao2016reducing}, image deblocking is modeled as an optimization problem within maximum a posteriori framework, and an algorithm that contains structural sparse representation (SSR) prior and quantization constraint (QC) prior is proposed for compression artifacts removal. \cite{kim2019pseudo} introduces a pseudo-blind system, which removes compression artifacts by estimating the quality factor and then applying a network that is trained with a similar quality factor. For JPEG image compression artifacts reduction, Jin \textit{et. al.} \cite{jin2020dual} propose a dual-stream recursive residual network (STRRN) which consists of structure and texture streams for separately reducing the specific artifacts related to high-frequency or low-frequency image components. Fu \textit{et. al.} \cite{fu2021learning} propose an interpretable deep network to learn both pixel-level regressive prior and semantic-level discriminative prior. In \cite{fu2021model}, a model-driven deep unfolding method is introduced to remove the artifacts. 
These methods are effective for compression artifacts reduction. However, compression artifacts are not easily perceived in low-light images. As a side effect of image enhancement processing, amplified artifacts along with color distortion are hard to be removed by the algorithms mentioned above.

\section{The Proposed Method}
\label{method}
\subsection{Overview}



We define the compressed dark image space as $\mathcal{C}$ and the normal-light image space as $\mathcal{N}$. Then, we define images sampled from these two spaces as $c \in \mathcal{C}$ and $n \in \mathcal{N}$. The enhancement can be established as a mapping from $\mathcal{C}$ to $\mathcal{N}$. The proposed method is implemented in following steps.


First, the image spaces $\mathcal{C}$ and $\mathcal{N}$ are encoded into their latent spaces $\mathcal{L}_{\mathcal{C}}$ and $\mathcal{L}_{\mathcal{N}}$ respectively as:
\begin{equation}
    \label{eq:1}
    \mathcal{L}_{\mathcal{C}} = \mathcal{E}_{C}(\mathcal{C}), ~~~
    \mathcal{L}_{\mathcal{N}} = \mathcal{E}_{N}(\mathcal{N})
\end{equation}
where $\mathcal{E}_{C}$ and $\mathcal{E}_{N}$ denote the encoding operation for $\mathcal{C}$ and $\mathcal{N}$ respectively.

Second, we learn two decoders that reconstruct the input images from the latent features:
\begin{equation}
    \label{eq:2}
    \mathcal{C} = \mathcal{D}_{C}(\mathcal{L}_{\mathcal{C}}), ~~~
    \mathcal{N} = \mathcal{D}_{N}(\mathcal{L}_{\mathcal{N}})
\end{equation}
where $\mathcal{D}_{C}$ and $\mathcal{D}_{N}$ denote the decoders for $\mathcal{L}_{\mathcal{C}}$ and $\mathcal{L}_{\mathcal{N}}$ respectively.

Third, we perform enhancement in the latent spaces. Specifically, we learn the latent mapping from $\mathcal{L}_{\mathcal{C}}$ to $\mathcal{L}_{\mathcal{N}}$ as:
\begin{equation}
    \label{eq:3}
    \mathcal{L}_{\mathcal{N}} = \mathcal{M}(\mathcal{L}_{\mathcal{C}})
\end{equation}
where $\mathcal{M}$ is the latent mapping operation.

By learning such a latent mapping operation, the whole enhancement can be formulated as:
\begin{equation}
    \label{eq:4}
    \mathcal{N} = \mathcal{D}_{N}(\mathcal{M}(\mathcal{E}_{C}(C)))
\end{equation}

Therefore, our approach consists of two stages: Firstly, we learn latent space $\mathcal{L}_{\mathcal{C}}$ and $\mathcal{L}_{\mathcal{N}}$ for compressed dark images and normal-light images respectively; Secondly, we learn latent space mapping function $\mathcal{M}$ to transform features from $\mathcal{L}_{\mathcal{C}}$ to $\mathcal{L}_{\mathcal{N}}$.

\subsection{Expand to Multiple Latent Spaces}
\begin{figure}[ht]
    \centering
    \includegraphics[scale=0.4]{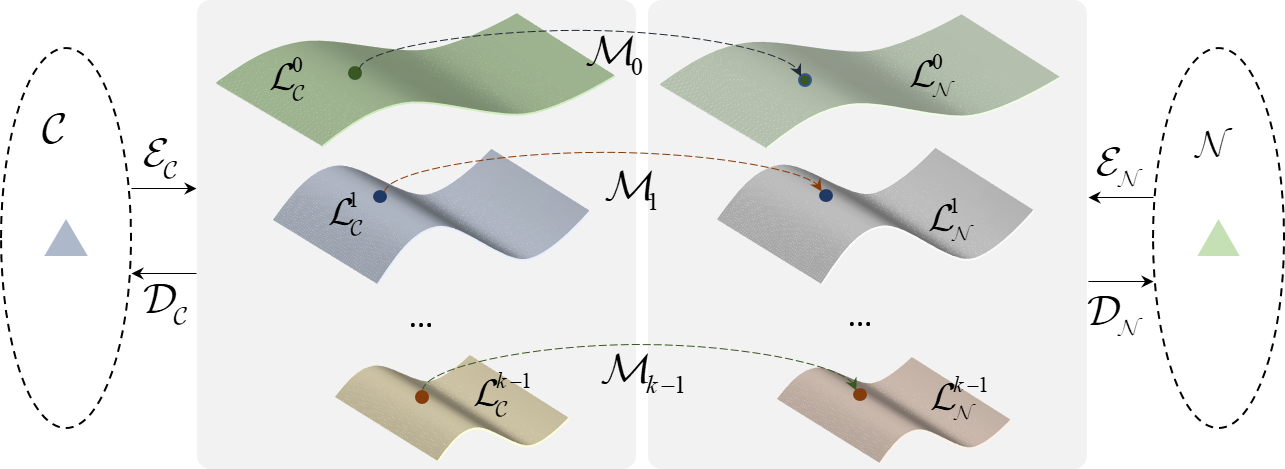}
    \caption{\textbf{Intuition of multiple latent space mapping.}}
    \label{fig:overview}
\end{figure}

Low-resolution features are rich in semantic information, and high-resolution features are rich in detailed information. The current single-level mapping model is difficult to consider both. In order to get enhanced results with high fidelity and rich textures, we expand the latent space to multi-level. The intuition of the proposed multiple latent space mapping is shown in Fig. \ref{fig:overview}.

First, we learn multiple latent spaces with multi-resolution latent features for $\mathcal{C}$ and $\mathcal{N}$ respectively. Therefore, Eq. \ref{eq:1} can be rewritten as:
\begin{equation}
    \label{eq:1-multi}
    \begin{aligned}
    &\mathcal{L}_{\mathcal{C}}^0,~ \mathcal{L}_{\mathcal{C}}^1,...,~ \mathcal{L}_{\mathcal{C}}^{k-1} = \mathcal{E}_{C}(\mathcal{C}) \\
    &\mathcal{L}_{\mathcal{N}}^0,~ \mathcal{L}_{\mathcal{N}}^1,...,~ \mathcal{L}_{\mathcal{N}}^{k-1} = \mathcal{E}_{N}(\mathcal{N})
    \end{aligned}
\end{equation}
where $\mathcal{E}_{C}$ and $\mathcal{E}_{N}$ denote the encoding operation for $\mathcal{C}$ and $\mathcal{N}$ respectively, $\{\mathcal{L}_{\mathcal{C}}^{i}\}_{i=0}^{k-1}$ and $\{\mathcal{L}_{\mathcal{N}}^{i}\}_{i=0}^{k-1}$ are the encoded multiple latent spaces which include multi-resolution latent features for $\mathcal{C}$ and $\mathcal{N}$ respectively. The feature resolution of $\{\mathcal{L}_{\mathcal{C}}^{i}\}_{i=0}^{k-1}$ is monotonically decreasing, so does $\{\mathcal{L}_{\mathcal{N}}^{i}\}_{i=0}^{k-1}$.

Second, we learn two decoders that reconstruct the multi-resolution latent features to the input images. Therefore, Eq. \ref{eq:2} can be rewritten as:
\begin{equation}
    \label{eq:2-multi}
    \begin{aligned}
    &\mathcal{C} = \mathcal{D}_{C}(\mathcal{L}_{\mathcal{C}}^0,~ \mathcal{L}_{\mathcal{C}}^1,...,~ \mathcal{L}_{\mathcal{C}}^{k-1}) \\
    &\mathcal{N} = \mathcal{D}_{N}(\mathcal{L}_{\mathcal{N}}^0,~ \mathcal{L}_{\mathcal{N}}^1,...,~ \mathcal{L}_{\mathcal{N}}^{k-1})
    \end{aligned}
\end{equation}
where $\mathcal{D}_{C}$ and $\mathcal{D}_{N}$ denote the decoders for $\{\mathcal{L}_{\mathcal{C}}^{i}\}_{i=0}^{k-1}$ and $\{\mathcal{L}_{\mathcal{N}}^{i}\}_{i=0}^{k-1}$ respectively.

Third, we learn multiple latent space mapping from $\{\mathcal{L}_{\mathcal{C}}^{i}\}_{i=0}^{k-1}$ to $\{\mathcal{L}_{\mathcal{N}}^{i}\}_{i=0}^{k-1}$. Therefore, Eq. \ref{eq:3} can be rewritten as:
\begin{equation}
    \label{eq:3-multi}
    \begin{aligned}
    & \mathcal{L}_{\mathcal{N}}^{k-1} = \mathcal{M}_{k-1}(\mathcal{L}_{\mathcal{C}}^{k-1}) \\
    & \mathcal{L}_{\mathcal{N}}^{k-2} = \mathcal{M}_{k-2}(\mathcal{L}_{\mathcal{C}}^{k-2}) \\
    & ~~~~~~~...... \\
    & \mathcal{L}_{\mathcal{N}}^0 = \mathcal{M}_0(\mathcal{L}_{\mathcal{C}}^0) \\
    \end{aligned}
\end{equation}

Finally, enhancement towards multiple latent space mapping can be formulated as:
\begin{equation}
    \label{eq:4}
    \mathcal{N} = \mathcal{D}_{N}(\{\mathcal{M}_0, \mathcal{M}_1, ..., \mathcal{M}_{k-1}\}(\mathcal{E}_{C}(C)))
\end{equation}


Next, we introduce the details of these two parts in Section \ref{vae} and Section \ref{mapping} respectively.

\subsection{Learning Multiple Latent Spaces}
\label{vae}
Since variational auto-encoder (VAE) \cite{kingma2013auto} encodes the input images to latent features with Gaussian distribution where the images can be reconstructed by sampling codes from latent space, we propose to learn multiple latent spaces for compressed dark images and normal-light images respectively by training two multi-level VAEs.
\begin{figure*}[ht]
    \centering
    \includegraphics[scale=0.55]{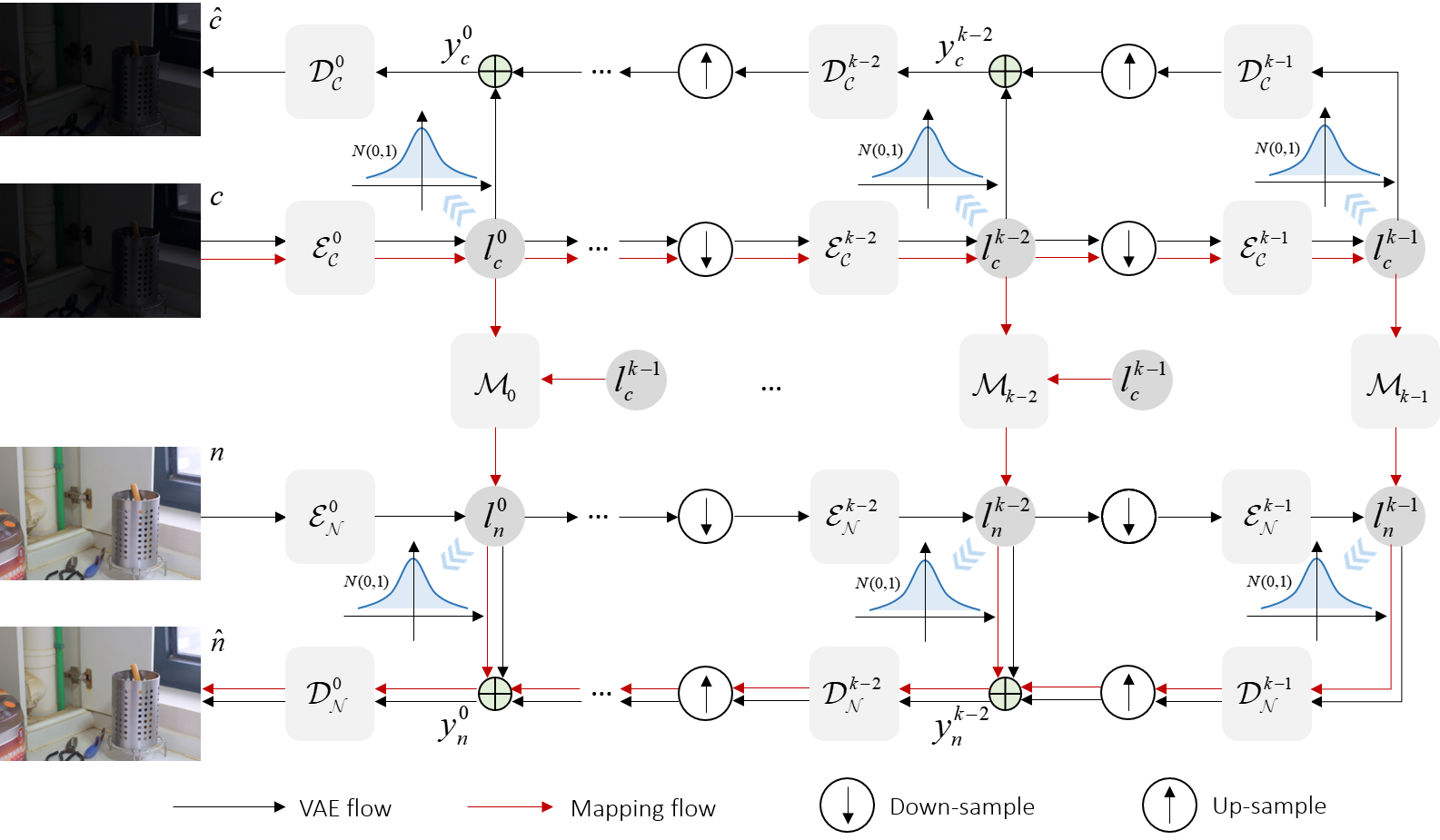}
    \caption{\textbf{Framework of the proposed multi-level VAEs and multiple latent space mapping.} }
    \label{fig:vaes}
\end{figure*}

\subsubsection{\textbf{Intuition of Multi-level VAE}}
The structure of our multi-level VAEs is shown in Fig. \ref{fig:vaes}. For compressed dark images, firstly, we use a group of sub-encoders to encode the input image to latent features as:
\begin{equation}
\label{encoder}
    \begin{aligned}
 & l_{c}^0 = \mathcal{E}_{\mathcal{C}}^{0}(c) \\
 & l_{c}^1 = \mathcal{E}_{\mathcal{C}}^{1}(l_{c}^{0}\downarrow) \\
 & ~~~~...... \\
 & l_{c}^{k-1} = \mathcal{E}_{\mathcal{C}}^{k-1}(l_{c}^{k-1}\downarrow) \\
    \end{aligned}
\end{equation}
where $\{l_{c}^i\}_{i=0}^{k-1}$ are the encoded latent features for the input compressed dark image $c$, $\{\mathcal{E}_{\mathcal{C}}^i\}_{i=0}^{k-1}$ are the sub-encoders, and $\downarrow$ means downsample by a factor of 2.

Then, we feed the encoded latent feature to a group of sub-decoders to reconstruct the input image as:
\begin{equation}
\label{decoder}
\begin{aligned}
 & y_{c}^{k-2} = \mathcal{D}_{\mathcal{C}}^{k-1}(l_{c}^{k-1})\uparrow + l_{c}^{k-2} \\
 & y_{c}^{k-3} = \mathcal{D}_{\mathcal{C}}^{k-2}(y_{c}^{k-2})\uparrow + l_{c}^{k-3} \\
 & ~~~~...... \\
 & \hat{c} = \mathcal{D}_{\mathcal{C}}^{0}(y_{c}^{0})
\end{aligned}
\end{equation}
where $\hat{c}$ is the reconstructed image of $c$, $\{\mathcal{D}_{\mathcal{C}}^i\}_{i=0}^{k-1}$ are the sub-decoders, and $\uparrow$ means upsample by a factor of 2.

Similarly, the reconstruction process for normal-light images is shown as:
\begin{equation}
\label{vae-n}
\begin{aligned}
& \{l_{n}^{i}\}_{i=0}^{k-1} = \{\mathcal{E}_{\mathcal{N}}^{i}\}_{i=0}^{k-1}(n) \\
& \hat{n} = \{\mathcal{D}_{\mathcal{N}}^{i}\}_{i=0}^{k-1}(\{l_{n}^{i}\}_{i=0}^{k-1})
\end{aligned}
\end{equation}
where $\{l_{n}^{i}\}_{i=0}^{k-1}$ are latent features for the input normal-light image $n$, $\hat{n}$ is the reconstructed image of $n$, $\{\mathcal{E}_{\mathcal{N}}^i\}_{i=0}^{k-1}$ are the sub-encoders for $n$, $\{\mathcal{D}_{\mathcal{N}}^i\}_{i=0}^{k-1}$ are the sub-decoders for $\{l_{n}^{i}\}_{i=0}^{k-1}$.

For realization, the sub-encoders $\{\mathcal{E}_{\mathcal{C}}^{i}\}_{i=0}^{k-1}$ and $\{\mathcal{E}_{\mathcal{N}}^{i}\}_{i=0}^{k-1}$ are realized by $3\times 3$ convolution followed by instance normalization and ReLU. The downsampling operation is realized by convolution with stride 2, and upsampling operation is realized by bilinear interpolation with scale factor of 2.

\begin{algorithm}[ht]
\caption{Learning Multiple Latent Spaces}
\begin{algorithmic}
\STATE 
\STATE \textbf{Require:} Functions $\{\mathcal{E}_{c}^i\}_{i=0}^{k-1}$ and $\{\mathcal{D}_{c}^i\}_{i=0}^{k-1}$, $c$ (batch of input images)

\STATE \textbf{1) Encode with Eq. \ref{encoder}}
\STATE \hspace{0.5cm} For $i\in \{0,1,...,k-1\}$:
\STATE \hspace{1cm} if $i==0$:
\STATE \hspace{1.5cm} $l_{c}^{i} \leftarrow \mathcal{E}_{\mathcal{C}}^{i}(c)$
\STATE \hspace{1cm} else:
\STATE \hspace{1.5cm} $l_{c}^{i} \leftarrow \mathcal{E}_{\mathcal{C}}^{i}(l_{c}^{i-1}\downarrow)$

\STATE \textbf{2) Decode with Eq. \ref{decoder}}
\STATE \hspace{0.5cm} For $i\in \{k-1,k-2,...,1\}$:
\STATE \hspace{1cm} if $i==k-1$:
\STATE \hspace{1.5cm} $y_{c}^{i-1} \leftarrow \mathcal{D}_{\mathcal{C}}^{i}(l_{c}^{i})\uparrow + l_{c}^{i-1}$
\STATE \hspace{1cm} else:
\STATE \hspace{1.5cm} $y_{c}^{i-1} \leftarrow \mathcal{D}_{\mathcal{C}}^{i}(y_{c}^{i})\uparrow + l_{c}^{i-1}$

\STATE \hspace{0.5cm} $\hat{c} \leftarrow \mathcal{D}_{C}^{0}(y_{c}^{0})$

\STATE \textbf{Loss Functions with Eq. \ref{eq:4-vaec-loss}}
\STATE \hspace{0.5cm} $\theta \leftarrow Update(L_{{\rm VAE_{\mathcal{C}}}}(c))$
\STATE Obtain latent features $\{l_{c}^{i}\}_{i=0}^{k-1}$ finally.
\end{algorithmic}
\label{alg1}
\end{algorithm}

\subsubsection{\textbf{Multi-level VAE Training}}
Our approach learns the multiple latent spaces with standard Gaussian distribution. During the training stage, the loss function of training VAE for compressed dark images is defined as:
\begin{equation}
    \label{eq:4-vaec-loss}
    \begin{aligned}
    L_{{\rm VAE_{\mathcal{C}}}}(c) = & ||\hat{c} - c|| + L_{{\rm PL}}(\hat{c}, c) + L_{{\rm VAE_{\mathcal{C}}, GAN}}(\hat{c}) \\
    & + \sum_{i=0}^{k-1}{\rm KL}(l_{c}^i, N(0, 1))
    \end{aligned}
\end{equation}
where $\{l_{c}^{i}\}_{i=0}^{k-1}$ are the multi-level latent features of the input image $c$, $\hat{c}$ is the reconstruction of $c$, and $N(0, 1)$ denotes the standard Gaussian distribution. The first item penalizes the least absolute deviations between $\hat{c}$ and $c$, which constrains the VAE to reconstruct the input image, so as to enforce latent
features to capture the major cues of images. The second item is perceptual loss \cite{johnson2016perceptual} that constrains the consistency of both high-level and low-level features. The third item is the least-square loss (LSGAN) \cite{mao2017least}, which encourages the VAE to reconstruct images with high realism. The fourth item is the
KL-divergence that constrains the multi-level latent features to squeeze to the standard Gaussian distribution $N(0, 1)$. 

Similar to Eq. \ref{eq:4-vaec-loss}, the loss function of training VAE$_{\mathcal{N}}$ can also be defined as:
\begin{equation}
    \label{eq:4-vaen-loss}
    \begin{aligned}
    L_{{\rm VAE_{\mathcal{N}}}}(n) = & ||\hat{n} - n|| + L_{{\rm PL}}(\hat{n}, n) + L_{{\rm VAE_{\mathcal{N}}, GAN}}(\hat{n}) \\
    & + \sum_{i=0}^{k-1}{\rm KL}(l_{n}^i, N(0, 1))
    \end{aligned}
\end{equation}


\subsection{Learning Latent Space Mapping}
\label{mapping}
The multi-level latent features are obtained after training two VAEs for compressed dark images and normal-light images respectively. We propose to learn the image enhancement by learning multiple latent space mapping. Such a mapping manner brings the benefit in two folds: 1) the mapping in the low-dimensional latent space is easier than that in the high-dimensional image space, 2) the multiple mapping mechanism preserves multi-scale information, and reduces position and detail information blur caused by downsampling in current single-level mapping.

\begin{algorithm}[ht]
\caption{Learning Latent Space Mapping Network}
\begin{algorithmic}
\STATE 
\STATE \textbf{Require:} Functions $\{\mathcal{E}_{c}^i\}_{i=0}^{k-1}$, $\{\mathcal{E}_{N}^i\}_{i=0}^{k-1}$, $\{\mathcal{D}_{N}^i\}_{i=0}^{k-1}$, and $\{\mathcal{M}\}_{i=0}^{k-1}$, $c$ (batch of compressed dark images), $n$ (batch of normal-light images)

\STATE \textbf{1) Encode $c$ with Eq. \ref{encoder}}
\STATE \hspace{0.5cm} $\{l_{c}^{i}\}_{i=0}^{k-1} \leftarrow \{\mathcal{E}_{\mathcal{C}}^{i}\}_{i=0}^{k-1}(c)$

\STATE \textbf{2) Encode $n$ with Eq. \ref{encoder}}
\STATE \hspace{0.5cm} $\{l_{n}^{i}\}_{i=0}^{k-1} \leftarrow \{\mathcal{E}_{\mathcal{N}}^{i}\}_{i=0}^{k-1}(n)$

\STATE \textbf{3) Map $\{l_{c}^{i}\}_{i=0}^{k-1}$ to $\{\widetilde{l}_{n}^{i}\}_{i=0}^{k-1}$}
\STATE \hspace{0.5cm} For $i\in \{k-1,k-2,...,0\}$:
\STATE \hspace{1cm} $\widetilde{l}_{n}^{i} \leftarrow \mathcal{M}_{i}(l_{c}^{i},~ l_{c}^{k-1}\uparrow)$

\STATE \textbf{4) Decode $\{\widetilde{l}_{n}^{i}\}_{i=0}^{k-1}$ with Eq. \ref{decoder}}
\STATE \hspace{0.5cm} $\widetilde{n} \leftarrow \{\mathcal{D}_{\mathcal{N}}^{i}\}_{i=0}^{k-1}(\{\widetilde{l}_{n}^{i}\}_{i=0}^{k-1})$
\STATE \textbf{Training Stage}
\STATE \hspace{0.5cm} Freeze $\{\mathcal{E}_{c}^i\}_{i=0}^{k-1}$, $\{\mathcal{E}_{N}^i\}_{i=0}^{k-1}$ and $\{\mathcal{D}_{N}^i\}_{i=0}^{k-1}$
\STATE \hspace{0.5cm} $\theta \leftarrow Update(L_{\mathcal{M}}(n, \widetilde{n}, \{l_{n}^{i}\}_{i=0}^{k-1}, \{\widetilde{l}_{n}^{i}\}_{i=0}^{k-1}))$
\STATE \textbf{Inference Stage}
\STATE \hspace{0.5cm} \textbf{Return} $\widetilde{n}$

\end{algorithmic}
\label{alg2}
\end{algorithm}

\begin{figure}[ht]
    \centering
    \includegraphics[scale=0.32]{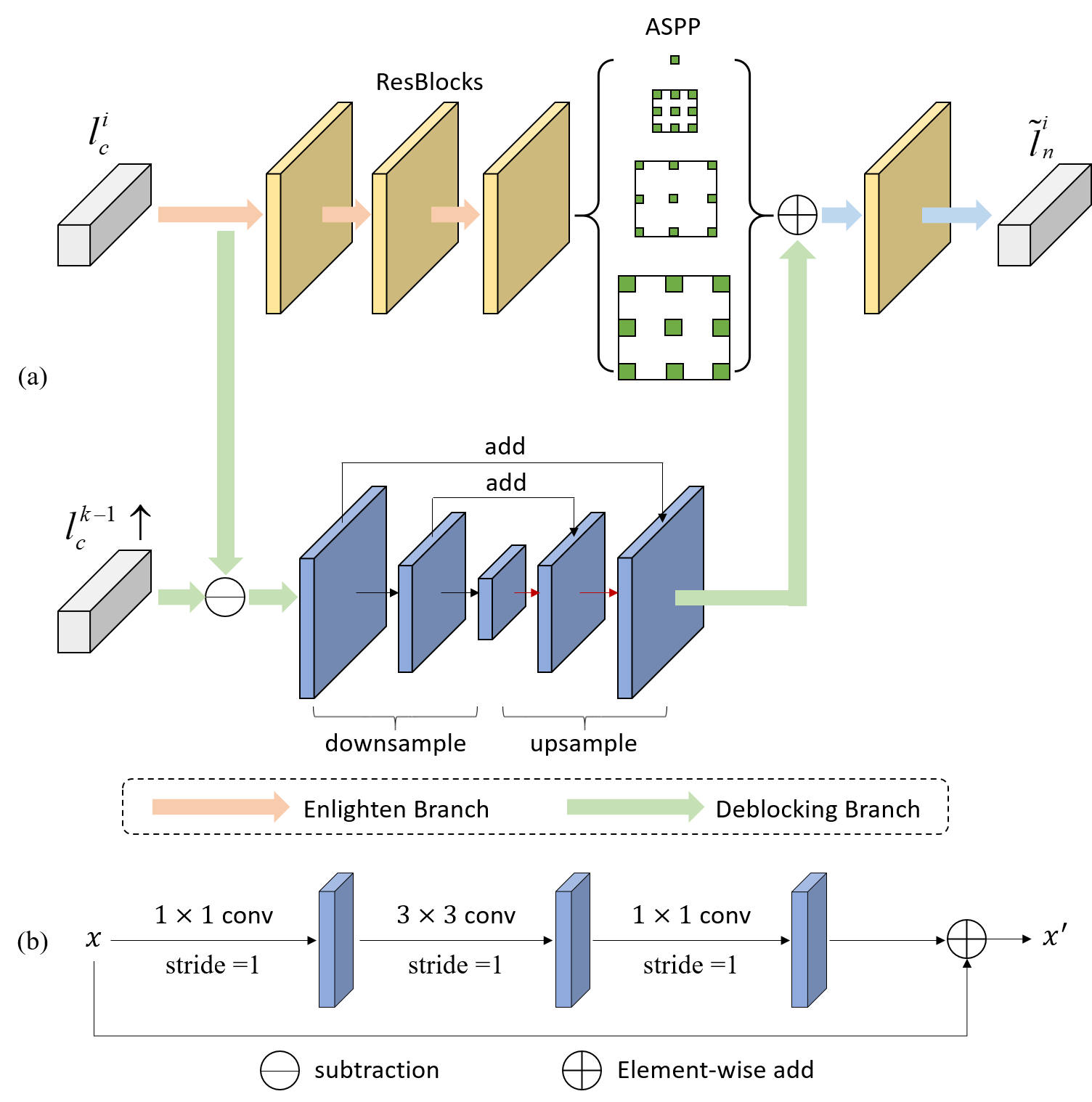}
    \caption{\textbf{Structure of latent space mapping network.} (a) the major part of latent space mapping network. (b) the ResBlock in (a). $l_{c}^{i}$ is the $i$-th level latent feature of compressed image. For enlighten branch, ASPP means Atrous Spatial Pyramid Pooling with different dilation rates as [1, 6, 12, 18]. For deblocking branch, the UNet-style network contains two downsample blocks and two upsample blocks. Each downsample block is realized by $3\times 3$ convolution with stride of 2 and padding of 1. Each upsample block is realized by deconvolution layers with scale factor of 2.}
    \label{fig:mapnet}
\end{figure}

\begin{figure*}
    \centering
    \includegraphics[scale=1]{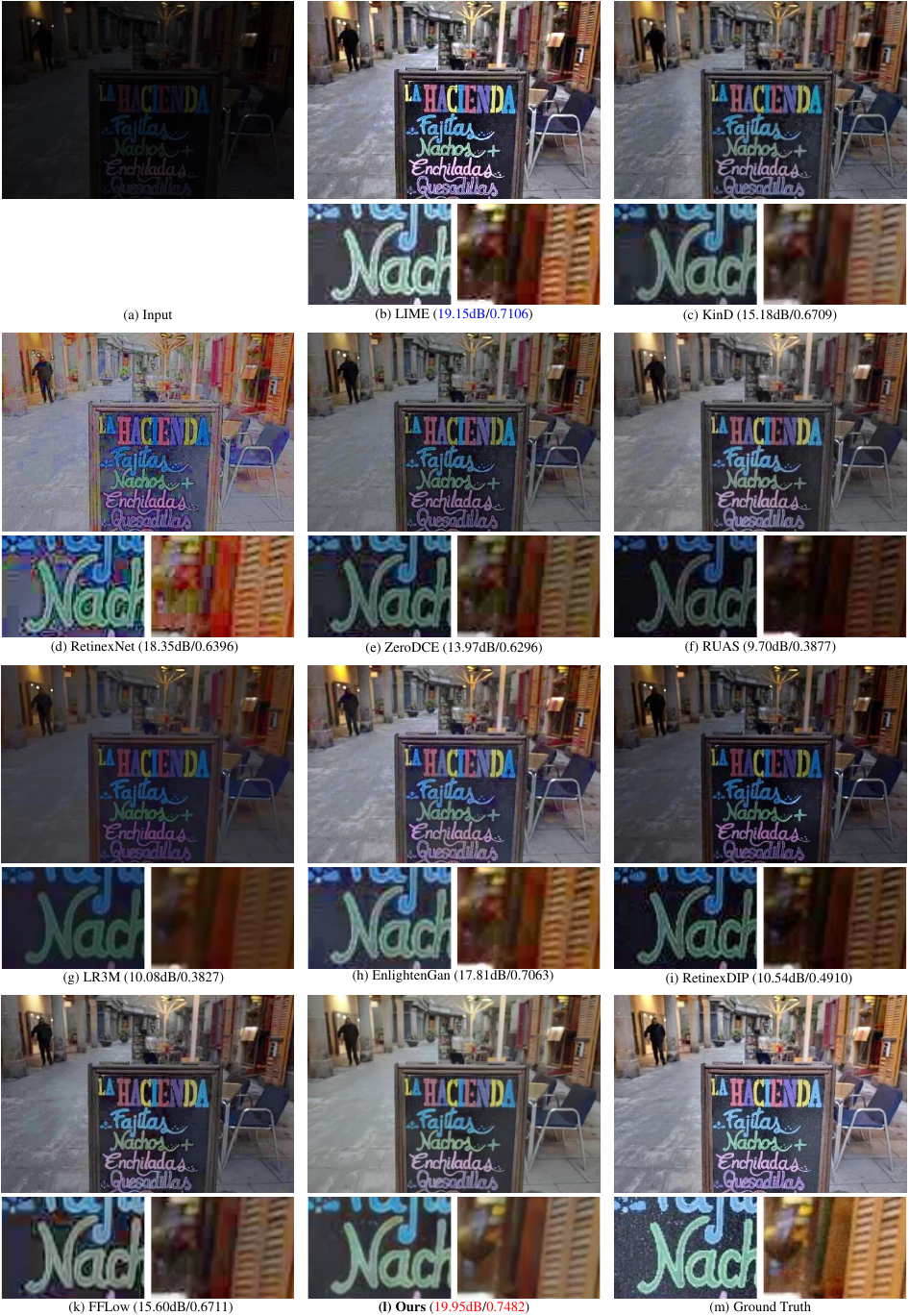}
    \caption{\textbf{Quantitative (PSNR/SSIM) and visual comparison with enhancement methods.} The partial details are displayed with zoom-in. The best metrics are in red and the second best are in blue.}
    \label{fig:enhance_compare}
\end{figure*}

\subsubsection{\textbf{Latent Space Mapping Network}}
Since the degradation models of darkness and compression are different from each other, we divide the latent space mapping into two branches, \textit{i.e.,} enlighten branch and deblocking branch. Specifically, we design the latent mapping network as Fig. \ref{fig:mapnet}.

\textbf{Enlighten Branch: }
Enlighten branch is used to restore the illumination cues and major details. Specifically, the enlighten branch takes the $i$-th latent feature $l_{c}^{i}$ as input. It firstly enters into two ResBlocks (shown in Fig. \ref{fig:mapnet} (b)). Then, the ASPP \cite{chen2017deeplab} (with dilation rate [1, 6, 12, 18]) is adopted to enhance the feature by aggregating multi-scale context cues. Finally, a ResBlock is leveraged to further enhance the feature with multi-scale context cues. The enlightening process is presented as:
\begin{equation}
    \label{eq:enlighten}
    F_{en} = f(l_{c}^{i})
\end{equation}
where $F_{en}$ denotes the output of enlighten branch, $f(\cdot)$ means the learnable function of enlighten branch.

\textbf{Deblocking Branch: }
Based on the fact that high-level features contain rich semantic cues and lack detail cues, we design the deblocking branch by considering the residual between $l_{c}^{i}$ and top-level feature $l_{c}^{k-1}$, as $l_{c}^{k-1}$ can be viewed as non-blocking feature (blocking artifact is a type of details). Therefore, the residual between $l_{c}^{i}$ and $l_{c}^{k-1}$ represents block-aware cues. Specifically, the deblocking branch firstly upsample $l_{c}^{k-1}$ to the size of $l_{c}^{i}$ as $l_{c}^{k-1} \uparrow$, and then compute the residual between $l_{c}^{i}$ and $l_{c}^{k-1} \uparrow$. Then, a UNet-style network is adopted to remove the blocking artifact cues. The deblocking process is presented as:
\begin{equation}
    \label{eq:deblocking}
    F_{de} = g(l_{c}^{i} - l_{c}^{k-1} \uparrow)
\end{equation}
where $F_{de}$ denotes the output of deblocking branch, $g(\cdot)$ means the learnable function of deblocking branch.

Finally, the latent space mapping network performs element-wise add between the outputs of enlightening branch and deblocking branch, and feeds it into a ResBlock to obtain the mapping result $\widetilde{l}_{n}^{i}$.

\subsubsection{\textbf{Intuition of Latent Space Mapping}}
Specifically, for the encoded latent features $\{l_{c}^{i}\}_{i=0}^{k-1}$, we propose to train $k$ latent space mapping networks as:
\begin{equation}
    \label{eq:9}
    \begin{aligned}
    & \widetilde{l}_{n}^{k-1} = \mathcal{M}_{k-1}(l_{c}^{k-1}, l_{c}^{k-1}\uparrow) \\
    & \widetilde{l}_{n}^{k-2} = \mathcal{M}_{k-2}(l_{c}^{k-1}, l_{c}^{k-1}\uparrow) \\
    & ~~~~~ ...... \\
    & \widetilde{l}_{n}^{0} = \mathcal{M}_0(l_{c}^{0}, l_{c}^{k-1}\uparrow) \\
    \end{aligned}
\end{equation}
where $\{\widetilde{l}_{n}^{i}\}_{i=0}^{k-1}$ are the predicted latent features for normal-light images whose ground-truths are $\{l_{n}^{i}\}_{i=0}^{k-1}$, and $\{\mathcal{M}_{i}\}_{i=0}^{k-1}$ are the mapping networks with the same structure.

Afterwards, the enhanced image $\widetilde{n}$ can be obtained by feeding $\{\widetilde{l}_{n}^{i}\}_{i=0}^{k-1}$ into $\{\mathcal{D}_{\mathcal{N}}^{i}\}_{i=0}^{k-1}$ based on Eq. \ref{decoder} as:
\begin{equation}
\label{decoder-mapping}
\begin{aligned}
 & y_{n}^{k-2} = \mathcal{D}_{\mathcal{N}}^{k-1}(\widetilde{l}_{n}^{k-1})\uparrow + l_{n}^{k-2} \\
 & y_{n}^{k-3} = \mathcal{D}_{\mathcal{N}}^{k-2}(y_{n}^{k-2})\uparrow + \widetilde{l}_{n}^{k-3} \\
 & ~~~~...... \\
 & \widetilde{n} = \mathcal{D}_{\mathcal{N}}^{0}(y_{n}^{0})
\end{aligned}
\end{equation}
For convenient, we simplify Eq.\ref{decoder-mapping} to $\widetilde{n} = \{\mathcal{D}_{\mathcal{N}}^{i}\}_{i=0}^{k-1}(\{\widetilde{l}_{n}^{i}\}_{i=0}^{k-1})$.

For realization, the structure of the mapping network is shown in Fig. \ref{fig:mapnet} (a). Firstly, $l_{c}^{'}$ is taken into two residual blocks whose structure is shown in Fig. \ref{fig:mapnet} (b). Then, channel attention and spatial attention are performed on the output of residual blocks. Finally, two residual blocks are taken to output the mapping latent feature $\widetilde{l}_{n}^{i}$.

\subsubsection{\textbf{Latent Space Mapping Training}}
When training latent space mapping networks, the parameters of two VAEs are frozen, and only the parameters of mapping networks can be updated. The loss function of training the latent mapping networks is defined as:
\begin{equation}
    \label{eq:11}
    L_{\mathcal{M}} = \sum_{i=0}^{k-1}||\widetilde{l}_{n}^{i} - l_{n}^{i}|| + L_{{\rm PL}}(\widetilde{n}, n) + L_{\rm LSGAN}(\widetilde{n})
\end{equation}
The first item penalizes the least absolute deviations between $\{\widetilde{l}_{n}^{i}\}_{i=0}^{k-1}$ and $\{l_{n}^i\}_{i=0}^{k-1}$, which enforce the networks to learn the mapping between compressed dark domain and normal-light domain. The second item is perceptual loss that constrains the consistency of both high-level and low-level features, so as to further enforce the enhancer to learn the enhancement process. The third item is the least-square loss (LSGAN) \cite{mao2017least} which encourages the ultimate enhancement results to look real.

\subsubsection{\textbf{The Whole Enhancement}}
After training two VAEs and the latent mapping networks, given a compressed dark image $c$, the enhancement result $\widetilde{n}$ can be obtained by:
\begin{equation}
    \label{eq:12}
    \widetilde{n} = \{\mathcal{D}_{\mathcal{N}}^{i}\}_{i=0}^{k-1} \circ \{\mathcal{M}\}_{i=0}^{k-1} \circ \{\mathcal{E}_{\mathcal{C}}^{i}\}_{i=0}^{k-1}(c)
\end{equation}

\begin{figure*}
    \centering
    \includegraphics[scale=1.23]{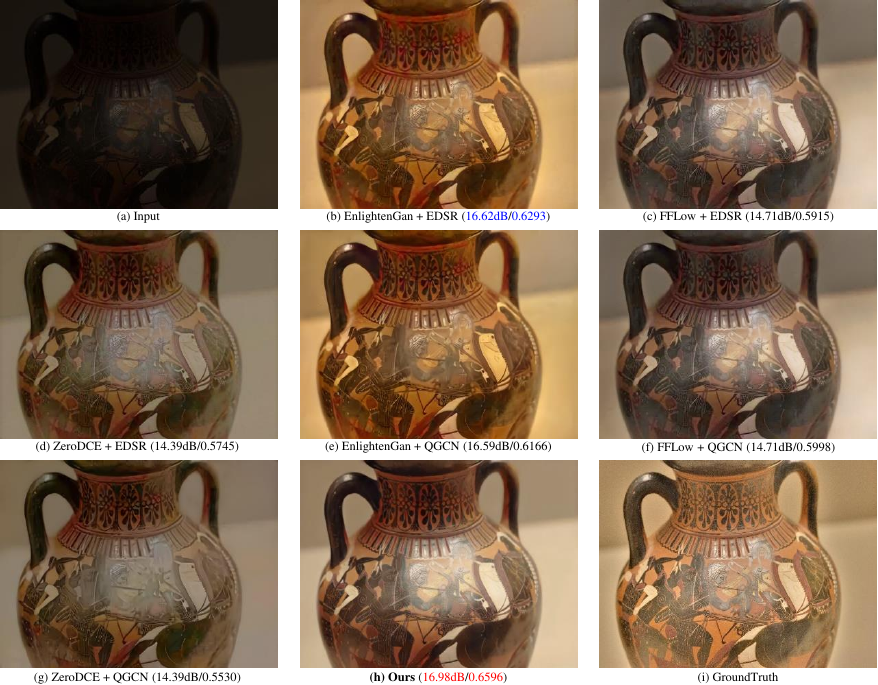}
    \caption{\textbf{Quantitative (PSNR/SSIM) and visual comparison   with enhancement+deblocking methods.} The best metrics are in red and the second best are in blue.}
    \label{fig:enhance_first}
\end{figure*}

\begin{figure*}
    \centering
    \includegraphics[scale=1.13]{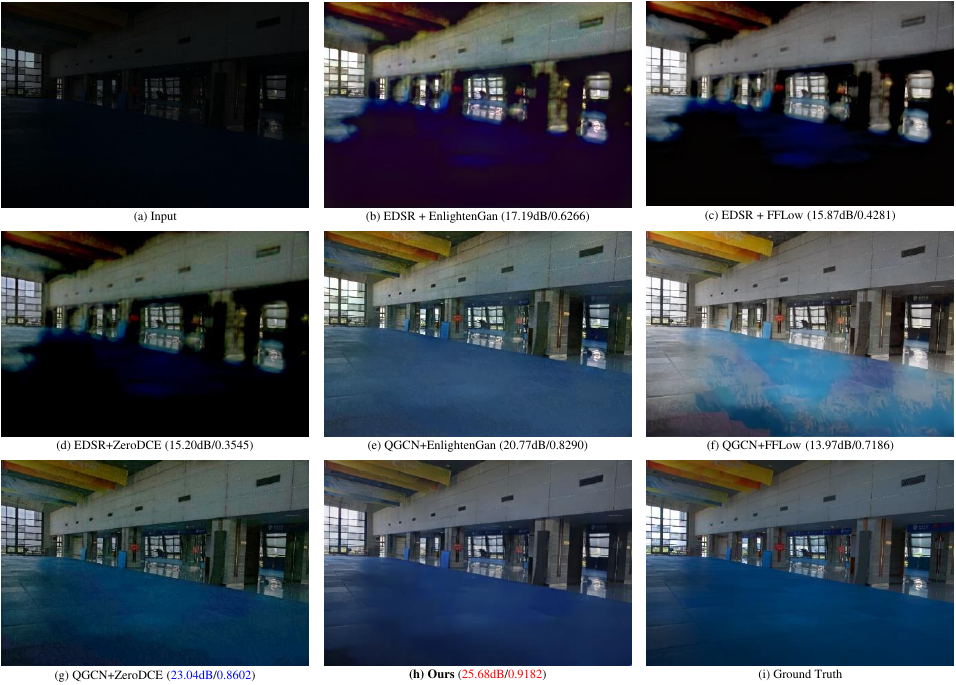}
    \caption{\textbf{Quantitative (PSNR/SSIM) and visual comparison with deblocking+enhancement methods.} The best metrics are in red and the second best are in blue.}
    \label{fig:deblock_first}
\end{figure*}

\section{Experiments}
\subsection{Implementation Details}
We implement the experiments on a Nvidia RTX 2080Ti GPU with PyTorch. We set the training batch size as 16. We utilize Adam optimizer for parameter learning. We set the start learning factor as 0.0002 and use linear learning factor decay. For each training iteration, we randomly crop patches of resolution $64 \times 64$. Our training process contains two stages: VAE stage and mapping stage. Firstly, we train two VAEs with 1000 epochs. Secondly, we freeze the parameters of two VAEs and train the latent mapping network with 500 epochs. We adopt $k=3$, \textit{i.e.,} three-level latent space mapping , as the default setting.


\subsection{Datasets}
To improve the robustness of the model, we merge several real-world datasets captured from different scenarios.

\emph{1) VAE Training Dataset:}
We project dark images and normal-light images to latent spaces through two VAEs, respectively. For the VAE of dark images, we collect 11831 dark images whose resources are exhibited in Table \ref{tab:vae-dark-set}. For the VAE of normal-light images, we use LabelME dataset \cite{russell2008labelme} which contains 36417 normal-light images, containing indoor, street, and natural scenarios \textit{et.al.}

\begin{table}[ht]
\centering
\caption{The resources of 10831 dark images for VAE. A7m3 means photos taken by ourselves with Sony A7m3 camera}
\setlength{\tabcolsep}{0.8mm}{
\scalebox{0.88}{
    \begin{tabular}{ccc}
    \toprule[0.7pt]
    Resource & Number & scenario\\ 
    \midrule[0.5pt]
    LOL \cite{Chen2018Retinex} & 485 & furniture \textit{et.al.}\\
    LSRW \cite{hai2021r2rnet} & 5600 & \makecell[c]{corridor, artificial constructions, food, dolls \textit{et.al.}}\\
    SICE \cite{cai2018learning} & 531 & \makecell[c]{natural scenarios, artificial constructions \textit{et.al.}}\\    
    RELLISUR \cite{aakerberg2021rellisur} & 3605 & \makecell[c]{papers with text, buildings, machinery \textit{et.al.}}\\
    A7m3 & 610 &  office, buildings \textit{et.al.} \\
    \bottomrule[0.7pt]
\end{tabular}
}}
\label{tab:vae-dark-set}
\end{table}

\begin{table}[ht]
\centering
\caption{The resources of 9923 image pairs for mapping training dataset. A7m3 means photos taken by ourselves with Sony A7m3 camera}
\setlength{\tabcolsep}{0.8mm}{
\scalebox{0.88}{
    \begin{tabular}{ccc}
    \toprule[0.7pt]
    Resource & Number & scenario\\ 
    \midrule[0.5pt]
    LOL \cite{Chen2018Retinex} & 485 & furnitures \textit{et.al.}\\
    LSRW \cite{hai2021r2rnet} & 5260 & corridor, artificial constructions, food, dolls \textit{et.al.}\\
    RELLISUR \cite{aakerberg2021rellisur} & 3568 & papers with text, buildings, machinery \textit{et.al.}\\
    A7m3 & 610 & office, buildings \textit{et.al.}\\
    \bottomrule[0.7pt]
\end{tabular}
}}
\label{tab:mapping-set}
\end{table}

\emph{2) Mapping Training Dataset:}
To train the latent mapping network, we collect 9923 image pairs whose resources are listed in Table \ref{tab:mapping-set}. Notely, since the dark images and reference images are misaligned in LSRW trainset and RELLSUR trainset, we adopt image registration algorithm\cite{bay2008speeded} to align them. Image pairs that fail to register are discarded.

\begin{table*}[ht]\Huge

\caption{\textbf{Average PSNR, SSIM and PSNR-B of enhanced results on LOL testset, LSRW testset and SICE dataset with QFs 90\%, 80\%.} The best results are in red and the second best are in blue.}
\centering
\tabcolsep=0.51cm
\scalebox{0.33}{
    \begin{tabular}{ccccccccccc}\toprule[2.5pt]
        \multirow{2}{*}{QF(\%)}&\multirow{2}{*}{Method}&\multicolumn{3}{c}{LOL}&\multicolumn{3}{c}{LSRW}&\multicolumn{3}{c}{SICE}\\
    \cmidrule(r){3-5}\cmidrule(r){6-8}\cmidrule(r){9-11} & &PSNR&SSIM&PSNR-B&PSNR&SSIM&PSNR-B&PSNR&SSIM&PSNR-B\\ \midrule[2.5pt]
    \multirow{21}{*}{90}& LIME\cite{guo2016lime}& 17.38 & 0.6353 & 17.16 & 14.89 & 0.4721 & 14.87 & 16.13 & 0.5030 & 16.25
    \\
     & RetinexNet\cite{Chen2018Retinex}& 16.28 & 0.5925 & 16.29 & 14.03 & 0.4118 & 14.06  & 16.01 & 0.4734 & 16.07
    \\
     & KinD\cite{zhang2019kindling} & 16.98 & 0.6997 & 17.08 & 14.97 & 0.4362 & 15.03 & 16.16 & 0.4918 & 15.67
    \\
     & EnlightenGAN\cite{jiang2021enlightengan} & 17.81 & 0.7404 & 17.73 & 15.75 & 0.4071 & 15.56 & 16.11 & 0.5071 & 16.28 
    \\
     & RetinexDIP\cite{zhao2021retinexdip} & 9.65 & 0.4316 & 9.76 & 11.74 & 0.2850 & 11.66  & 11.22 & 0.3507 & 11.60 
    \\ 
     & Zero-DCE\cite{guo2020zero} & 15.67 & 0.6217 & 15.67 & 14.93 & 0.3946 & 15.35 & 14.81 & 0.4771 & 15.53
     \\
     & LLFlow\cite{wang2021low} & 17.45 & 0.7633 & 17.59  & 15.38 & 0.4314 & 15.41 & 15.07 & 0.5004 & 15.63
     \\
     & RUAS\cite{liu2021retinex} & 16.47 & 0.7271 & 16.41 & 12.12 & 0.4527 & 12.13 & 11.47 & 0.4271 & 11.41

    \\\cmidrule(r){2-11}
     & EnlightenGAN\cite{jiang2021enlightengan} + EDSR\cite{lim2017enhanced} & 17.98 & 0.7747 & 17.99 & 15.95 & 0.4360 & 15.79 & \textcolor[rgb]{1,0,0}{16.30} & 0.5131 & 16.31
    \\
     & ZeroDCE\cite{guo2020zero} + EDSR\cite{lim2017enhanced}  & 15.87 & 0.6599 & 15.89 & 15.15 & 0.4213 & 15.64 & 14.93 & 0.4783 & 15.63
    \\
    & LLFlow\cite{wang2021low} + EDSR\cite{lim2017enhanced} &  17.56 & 0.7884 & 17.70 & 15.53 & 0.4589 & 15.55 & 15.15 & 0.5048 & 15.62
    \\
     & EnlightenGAN\cite{jiang2021enlightengan} + QGCN\cite{li-2020}  & 17.83 & 0.7918 & 17.88 & \textcolor[rgb]{0,0,1}{16.07} & \textcolor[rgb]{0,0,1}{0.4839} & \textcolor[rgb]{0,0,1}{15.95} & 16.19 & 0.5021 & \textcolor[rgb]{0,0,1}{16.40}
    \\    
     & ZeroDCE\cite{guo2020zero} + QGCN\cite{li-2020}  & 16.92 & 0.7713 & 17.71 & 15.27 & 0.4805 & 15.82 & 14.92 & 0.4659 & 15.62
    \\
    & LLFlow\cite{wang2021low} + QGCN\cite{li-2020} &  17.51 & 0.7855 & 17.69 & 15.46 & 0.4562 & 15.51 & 15.14 & 0.5070 & 15.72
    \\\cmidrule(r){2-11}
     & EDSR\cite{lim2017enhanced} + EnlightenGAN\cite{jiang2021enlightengan} & 15.29 & 0.6311 & 15.67 & 15.68 & 0.4326 & 15.63 & 14.70 & 0.4703 & 14.87 
     \\
     & EDSR\cite{lim2017enhanced} + ZeroDCE\cite{guo2020zero} & 12.61 & 0.5138 & 12.41 & 14.71 & 0.3785 & 14.99 & 13.55 & 0.4215 & 13.91
     \\
     & EDSR\cite{lim2017enhanced} + LLFlow\cite{wang2021low} & 14.18 & 0.5689 & 14.01 & 14.49 & 0.4005 & 14.40 & 13.62 & 0.4367 & 14.00
     \\
     & QGCN\cite{li-2020} + EnlightenGAN\cite{jiang2021enlightengan} & 18.07 & 0.7985 & 17.97 & 15.91 & 0.4405 & 15.69  & 15.74 & \textcolor[rgb]{0,0,1}{0.5373} & 15.73 
     \\
     & QGCN\cite{li-2020} + ZeroDCE\cite{guo2020zero} & 14.97 & 0.7324 & 15.01 & 15.05 & 0.4340 & 15.48 & 14.66 & 0.5079 & 15.22 
     \\
     & QGCN\cite{li-2020} + LLFlow\cite{wang2021low} & \textcolor[rgb]{0,0,1}{18.27} & \textcolor[rgb]{0,0,1}{0.8113} & \textcolor[rgb]{0,0,1}{18.25} & 15.32 & 0.4617 & 15.37 & 15.52 & \textcolor[rgb]{1,0,0}{0.5452} & 16.08 
     \\\cmidrule(r){2-11}
     & \textbf{Ours} & \textcolor[rgb]{1,0,0}{\textbf{19.16}} & \textcolor[rgb]{1,0,0}{\textbf{0.8373}} & \textcolor[rgb]{1,0,0}{\textbf{19.06}} & \textcolor[rgb]{1,0,0}{\textbf{16.83}} & \textcolor[rgb]{1,0,0}{\textbf{0.4959}} & \textcolor[rgb]{1,0,0}{\textbf{16.81}} & \textcolor[rgb]{0,0,1}{\textbf{16.20}} & 0.5189 & \textcolor[rgb]{1,0,0}{\textbf{16.89}}
     \\
    \midrule[2.5pt]
    \multirow{21}{*}{80} & LIME\cite{guo2016lime}& 17.39 & 0.7035 & 17.21 & 14.43 & 0.3921 & 14.41 & \textcolor[rgb]{0,0,1}{16.13} & 0.5118 & \textcolor[rgb]{0,0,1}{16.21}
    \\
     & RetinexNet\cite{Chen2018Retinex}& 17.03 & 0.6734 & 16.84 & 13.44 & 0.3356 & 13.17 & 14.95 & 0.4625 & 14.98 
    \\
     & KinD\cite{zhang2019kindling} & 17.20 & 0.7400 & 17.25 & 15.28 & 0.4392 & 15.34 & 15.16 & 0.5045 & 15.51
    \\
     & EnlightenGAN\cite{jiang2021enlightengan} & 18.03 & 0.7620 & 17.94 & 15.94 & 0.4447 & 15.74 & 15.98 & 0.5164 & 16.09 
    \\
     & RetinexDIP\cite{zhao2021retinexdip} & 9.76 & 0.4188 & 10.16 & 12.59 & 0.3762 & 12.54 & 11.18 & 0.3549 & 11.57
    \\
     & Zero-DCE\cite{guo2020zero} & 14.99 & 0.7140 & 15.14 & 15.05 & 0.4409 & 15.46 & 14.73 & 0.4868 & 15.41
     \\
     & LLFlow\cite{wang2021low} & 17.01 & 0.7606 & 17.28 & 15.28 & 0.4573 & 15.30 & 15.19 & 0.5158 & 15.72
     \\
     & RUAS\cite{liu2021retinex} & 16.38 & 0.7195 & 16.25 & 13.76 & 0.3876 & 13.80 & 11.39 & 0.3710 & 12.15

    \\\cmidrule(r){2-11}
     & EnlightenGAN\cite{jiang2021enlightengan} + EDSR\cite{lim2017enhanced} & 18.23 & 0.7802 & \textcolor[rgb]{0,0,1}{18.27} & 16.13 & 0.4686 & 15.95 & 16.09 & 0.5235 & 16.14
    \\
     & ZeroDCE\cite{guo2020zero} + EDSR\cite{lim2017enhanced}  & 15.04 & 0.7266 & 15.20 & 15.23 & 0.4653 & 15.71 & 14.86 & 0.4882 & 15.53 
    \\
    & LLFlow\cite{wang2021low} + EDSR\cite{lim2017enhanced} & 17.12 & 0.7910 & 17.43 & 15.41 & \textcolor[rgb]{0,0,1}{0.4813} & 15.43 & 15.29 & 0.5212 & 15.72
    \\
     & EnlightenGAN\cite{jiang2021enlightengan} + QGCN\cite{li-2020}  & 18.07 & 0.7888 & 18.06 & \textcolor[rgb]{0,0,1}{16.15} & 0.4788 & \textcolor[rgb]{0,0,1}{15.98} & 16.05 & 0.4977 & 16.18
    \\    
     & ZeroDCE\cite{guo2020zero} + QGCN\cite{li-2020}  & 14.92 & 0.7157 & 15.05 & 15.24 & 0.4727 & 15.78 & 14.85 & 0.4640 & 15.52
    \\
    & LLFlow\cite{wang2021low} + QGCN\cite{li-2020} & 17.11 & 0.7864 & 17.45 & 15.32 & 0.4707 & 15.40 & 15.26 & 0.5217 & 15.80 
    \\\cmidrule(r){2-11}
     & EDSR\cite{lim2017enhanced} + EnlightenGAN\cite{jiang2021enlightengan} & 15.26 & 0.6298 & 15.61 & 15.66 & 0.4445 & 15.67 & 14.77 & 0.4627 & 14.94 
     \\
     & EDSR\cite{lim2017enhanced} + ZeroDCE\cite{guo2020zero} & 12.56 & 0.5109 & 12.32 & 14.72 & 0.3875 & 15.00 & 13.52 & 0.4109 & 13.85
     \\
     & EDSR\cite{lim2017enhanced} + LLFlow\cite{wang2021low} & 13.99 & 0.5647 & 13.83 & 14.42 & 0.4056 & 14.36 & 13.55 & 0.4277 & 13.93
     \\
     & QGCN\cite{li-2020} + EnlightenGAN\cite{jiang2021enlightengan} & \textcolor[rgb]{0,0,1}{18.25} & \textcolor[rgb]{0,0,1}{0.8124} & 18.21 & 15.93 & 0.4656 & 15.75 & 16.11 & \textcolor[rgb]{0,0,1}{0.5297} & 16.01
     \\
     & QGCN\cite{li-2020} + ZeroDCE\cite{guo2020zero} & 14.99 & 0.7353 & 14.99 & 15.11 & 0.4612 & 15.56 & 14.77 & 0.4927 & 15.29
     \\
     & QGCN\cite{li-2020} + LLFlow\cite{wang2021low} & 17.77 & 0.7985 & 17.86 & 15.14 & 0.4767 & 15.21 & 15.45 & 0.5249 & 16.06
     \\\cmidrule(r){2-11}
     & \textbf{Ours} & \textcolor[rgb]{1,0,0}{\textbf{18.92}} & \textcolor[rgb]{1,0,0}{\textbf{0.8159}} & \textcolor[rgb]{1,0,0}{\textbf{18.97}}& \textcolor[rgb]{1,0,0}{\textbf{16.54}} & \textcolor[rgb]{1,0,0}{\textbf{0.4947}} & \textcolor[rgb]{1,0,0}{\textbf{16.51}} & \textcolor[rgb]{1,0,0}{\textbf{16.59}} & \textcolor[rgb]{1,0,0}{\textbf{0.5314}} & \textcolor[rgb]{1,0,0}{\textbf{17.27}}
    \\
\bottomrule[2.5pt]
\end{tabular}}
\label{tab:compara-90-80}
\end{table*}

\begin{table*}\Huge

\caption{\textbf{Average PSNR, SSIM and PSNR-B and of enhanced results on LOL testset, LSRW testset and SICE dataset with QFs 70\%, 60\%.} The best results are in red and the second best are in blue.}
\centering
\tabcolsep=0.51cm
\scalebox{0.33}{
    \begin{tabular}{ccccccccccc}\toprule[2.5pt]
        \multirow{2}{*}{QF(\%)}&\multirow{2}{*}{Method}&\multicolumn{3}{c}{LOL}&\multicolumn{3}{c}{LSRW}&\multicolumn{3}{c}{SICE}\\
    \cmidrule(r){3-5}\cmidrule(r){6-8}\cmidrule(r){9-11} & &PSNR&SSIM&PSNR-B&PSNR&SSIM&PSNR-B&PSNR&SSIM&PSNR-B\\ \midrule[2.5pt]
    \multirow{21}{*}{70} & LIME\cite{guo2016lime}& 17.42 & 0.7089 & 17.35 & 14.46 & 0.4044 & 14.43 & \textcolor[rgb]{0,0,1}{16.12} & 0.5030 & 16.28
    \\
     & RetinexNet\cite{Chen2018Retinex}& 17.09 & 0.6828 & 16.96  & 13.48 & 0.3491 & 13.26 & 14.91 & 0.4503 & 15.01
    \\
     & KinD\cite{zhang2019kindling} & 17.28 & 0.7319 & 17.26 & 15.24 & 0.4427 & 15.27 & 15.08 & 0.4913 & 15.45
    \\
     & EnlightenGAN\cite{jiang2021enlightengan} & 18.21 & 0.7585 & 18.22 & 15.94 & 0.4505 & 15.73 & 15.12 & 0.5061 & 15.33 
    \\
     & RetinexDIP\cite{zhao2021retinexdip} & 9.24 & 0.3972 & 9.58 & 12.62 & 0.3872 & 12.61 & 11.20 & 0.3524 & 11.60
    \\
     & Zero-DCE\cite{guo2020zero} & 15.04 & 0.7108 & 15.27 & 15.09 & 0.4490 & 15.50 & 14.80 & 0.4794 & 15.54 
     \\
     & LLFlow\cite{wang2021low} & 16.90 & 0.7519 & 17.23 & 15.25 & 0.4609 & 15.27 & 15.10 & 0.5015 & 15.67 

     \\
     & RUAS\cite{liu2021retinex} & 16.39 & 0.7136 & 16.36 & 13.80 & 0.3938 & 13.84 & 11.40 & 0.3669 & 12.20

    \\\cmidrule(r){2-11}
     & EnligntenGAN\cite{jiang2021enlightengan} + EDSR\cite{lim2017enhanced} & \textcolor[rgb]{0,0,1}{18.44} & 0.7868 & 18.25 & \textcolor[rgb]{0,0,1}{16.11} & 0.4745 & \textcolor[rgb]{0,0,1}{15.93} & 15.22 & \textcolor[rgb]{0,0,1}{0.5139} & 15.38 
    \\
     & ZeroDCE\cite{guo2020zero} + EDSR\cite{lim2017enhanced}  & 15.10 & 0.7269 & 15.32 & 15.26 & 0.4736 & 15.73 & 14.91 & 0.4796 & 15.64
    \\
    & LLFlow\cite{wang2021low} + EDSR\cite{lim2017enhanced} &  17.03 & 0.7844 & 17.40 & 15.37 & \textcolor[rgb]{0,0,1}{0.4854} & 15.38 & 15.19 & 0.5075 & 15.67
    \\
     & EnligntenGAN\cite{jiang2021enlightengan} + QGCN\cite{li-2020}  & 18.29 & 0.7812 & \textcolor[rgb]{0,0,1}{18.31} & 16.07 & 0.4744 & 15.92 & 15.23 & 0.5067 & 15.48
    \\    
     & ZeroDCE\cite{guo2020zero} + QGCN\cite{li-2020} & 15.04 & 0.7234 & 15.28 & 15.23 & 0.4724 & 15.77 & 14.93 & 0.4729 & 15.66
     \\
    & LLFlow\cite{wang2021low} + QGCN\cite{li-2020} &  17.00 & 0.7731 & 17.39 & 15.28 & 0.4707 & 15.35 & 15.18 & 0.5095 & 15.77
    \\\cmidrule(r){2-11}
     & EDSR\cite{lim2017enhanced} + EnlightenGAN\cite{jiang2021enlightengan} & 15.36 & 0.6282 & 15.84 & 15.67 & 0.4477 & 15.69 & 14.82 & 0.4572 & 15.04
     \\
     & EDSR\cite{lim2017enhanced} + ZeroDCE\cite{guo2020zero} & 12.63 & 0.5106 & 12.44 & 14.74 & 0.3904 & 15.03 & 13.55 & 0.4059 & 13.91
     \\
     & EDSR\cite{lim2017enhanced} + LLFlow\cite{wang2021low} & 13.93 & 0.5615 & 13.82 & 14.42 & 0.4078 & 14.36 & 13.52 & 0.4219 & 13.94
     \\
     & QGCN\cite{li-2020} + EnlightenGAN\cite{jiang2021enlightengan} & 18.41 & \textcolor[rgb]{1,0,0}{0.8097} & 18.29 & 15.90 & 0.4727 & 15.72  & \textcolor[rgb]{1,0,0}{16.28} & \textcolor[rgb]{1,0,0}{0.5198} & \textcolor[rgb]{0,0,1}{16.32}
     \\
     & QGCN\cite{li-2020} + ZeroDCE\cite{guo2020zero} & 15.09 & 0.7331 & 15.21 & 15.16 & 0.4683 & 15.61 & 14.86 & 0.4848 & 15.47
     \\
     & QGCN\cite{li-2020} + LLFlow\cite{wang2021low} & 17.64 & 0.7990 & 17.84 & 15.15 & 0.4832 & 15.19 & 15.42 & 0.5144 & 16.07
     \\\cmidrule(r){2-11}
     & \textbf{Ours} & \textcolor[rgb]{1,0,0}{\textbf{18.46}} & \textcolor[rgb]{0,0,1}{\textbf{0.7991}} & \textcolor[rgb]{1,0,0}{\textbf{18.32}} &  \textcolor[rgb]{1,0,0}{\textbf{16.86}} & \textcolor[rgb]{1,0,0}{\textbf{0.4982}} & \textcolor[rgb]{1,0,0}{\textbf{16.86}} &  \textbf{15.90} & \textbf{0.5073} & \textcolor[rgb]{1,0,0}{\textbf{16.68}}
     \\
    \midrule[2.5pt]
    \multirow{21}{*}{60} & LIME\cite{guo2016lime}& 17.20 & 0.6939 & 17.00 & 14.45 & 0.4075 & 14.39 & 15.82 & 0.4853 & 15.78
    \\
     & RetinexNet\cite{Chen2018Retinex}& 16.76 & 0.6750 & 16.63  & 13.45 & 0.3559 & 13.19 &  14.58 & 0.4388 & 14.52 
    \\
     & KinD\cite{zhang2019kindling} & 17.15 & 0.7160 & 16.99 & 15.21 & 0.4411 & 15.27 & 14.76 & 0.4734 & 14.90
    \\
     & EnlightenGAN\cite{jiang2021enlightengan} & 18.13 & 0.7450 & 18.13 & 15.78 & 0.4514 & 15.54 & 15.05 & 0.4962 & 15.02
    \\
     & RetinexDIP\cite{zhao2021retinexdip} & 9.72 & 0.4250 & 10.03 & 12.58 & 0.3868 & 12.50 & 11.08 & 0.3351 & 11.46 
    \\
     & Zero-DCE\cite{guo2020zero} & 14.94 & 0.6942 & 15.05 & 15.05 & 0.4473 & 15.45 & 14.78 & 0.4650 & 15.36 
     \\
     & LLFlow\cite{wang2021low} & 16.74 & 0.7343 & 17.12 & 15.17 & 0.4587 & 15.17 & 15.08 & 0.4865 & 15.60
 
     \\
     & RUAS\cite{liu2021retinex} & 16.19 & 0.7006 & 15.97 & 13.79 & 0.3932 & 13.83 & 11.40 & 0.3601 & 12.14

    \\\cmidrule(r){2-11}
     & EnligntenGAN\cite{jiang2021enlightengan} + EDSR\cite{lim2017enhanced} & \textcolor[rgb]{0,0,1}{18.44} & 0.7778 & \textcolor[rgb]{1,0,0}{18.61} & \textcolor[rgb]{0,0,1}{15.96} & 0.4774 & \textcolor[rgb]{0,0,1}{15.76} & 15.15 & \textcolor[rgb]{0,0,1}{0.5048} & 15.07
    \\
     & ZeroDCE\cite{guo2020zero} + EDSR\cite{lim2017enhanced}  & 15.00 & 0.7092 & 15.14 & 15.23 & 0.4706 & 15.70 & 14.88 & 0.4664 & 15.45
    \\
    & LLFlow\cite{wang2021low} + EDSR\cite{lim2017enhanced} & 16.90 & 0.7716 & 17.35 & 15.31 & \textcolor[rgb]{0,0,1}{0.4856} & 15.31 & 15.18 & 0.4943 & 15.62
    \\
     & EnlightenGAN\cite{jiang2021enlightengan} + QGCN\cite{li-2020}  & 18.34 & 0.7811 & 18.37 & 15.90 & 0.4738 & 15.72 & 15.20 & 0.5055 & 15.27
    \\    
     & ZeroDCE\cite{guo2020zero} + QGCN\cite{li-2020}  & 15.08 & 0.7170 & 15.33 & 15.15 & 0.4684 & 15.62 & 14.92 & 0.4673 & 15.60
    \\
    & LLFlow\cite{wang2021low} + QGCN\cite{li-2020} &  16.83 & 0.7520 & 17.27 & 15.22 & 0.4690 & 15.26 & 15.17 & 0.4949 & 15.71
    \\\cmidrule(r){2-11}
     & EDSR\cite{lim2017enhanced} + EnlightenGAN\cite{jiang2021enlightengan} & 15.27 & 0.6209 & 15.58 & 15.53 & 0.4474 & 15.54 & 14.73 & 0.4519 & 14.83
     \\
     & EDSR\cite{lim2017enhanced} + ZeroDCE\cite{guo2020zero} & 12.56 & 0.5030 & 12.30 & 14.70 & 0.3886 & 14.99 & 13.53 & 0.3999 & 13.81
     \\
     & EDSR\cite{lim2017enhanced} + LLFlow\cite{wang2021low} & 17.38 & 0.7737 & 17.65 & 15.19 & 0.4827 & 15.24 & 15.33 & 0.4987 & 15.93
     \\
     & QGCN\cite{li-2020} + EnlightenGAN\cite{jiang2021enlightengan} & 18.33 & \textcolor[rgb]{1,0,0}{0.7970} & 18.42 & 15.75 & 0.4755 & 15.53 & \textcolor[rgb]{0,0,1}{16.19} & \textcolor[rgb]{1,0,0}{0.5112} & \textcolor[rgb]{0,0,1}{16.02}
     \\
     & QGCN\cite{li-2020} + ZeroDCE\cite{guo2020zero} & 15.00 & 0.7158 & 15.03 & 15.15 & 0.4673 & 15.63 & 14.85 & 0.4718 & 15.32
     \\
     & QGCN\cite{li-2020} + LLFlow\cite{wang2021low} & 17.38 & 0.7737 & 17.65 & 15.19 & 0.4827 & 15.24 & 15.33 & 0.4987 & 15.93
     \\\cmidrule(r){2-11}
     & \textbf{Ours} & \textcolor[rgb]{1,0,0}{\textbf{18.65}} & \textcolor[rgb]{0,0,1}{\textbf{0.7901}} & \textcolor[rgb]{0,0,1}{\textbf{18.46}} &  \textcolor[rgb]{1,0,0}{\textbf{17.12}} & \textcolor[rgb]{1,0,0}{\textbf{0.4966}} & \textcolor[rgb]{1,0,0}{\textbf{17.05}}& \textcolor[rgb]{1,0,0}{\textbf{16.45}} & \textbf{0.4992} & \textcolor[rgb]{1,0,0}{\textbf{17.01}}
    \\
\bottomrule[2.5pt]
\end{tabular}}
\label{tab:compara-70-60}
\end{table*}

\emph{3) Test Dataset:}
We use LOL testset, LSRW testset and SICE testset to test the performance of our approach.


\subsection{Evaluation Metrics}
Peak Signal-to-Noise Ratio (PSNR), Structure Similarity (SSIM) and PSNR-B \cite{yim2010quality} are widely used to evaluate the performance of compression artifacts reduction and light restoration. A higher PSNR value
indicates a test image is closer to the reference image in terms of pixel-level image content. A
higher SSIM value indicates a test image is closer to the reference image in terms of structural properties. PSNR-B is a block-sensitive image quality index designed to measure the blocking artifacts, and a higher PSNR-B value indicates less blocking artifacts in test images.

\subsection{Qualitative Evaluation}
We present the visual comparisons on typical compressed dark images, and compare our approach with \textit{i.e.,} \textbf{enhancement} methods, \textbf{enhancement+deblocking} methods and \textbf{deblocking+enhancement} methods.

\subsubsection{Compare with \textbf{Enhancement Methods}}

We show comparative visualization results in Fig. \ref{fig:enhance_compare} to intuitively show performance. It is seen that most previous enhancement methods efficiently improve image brightness and contrast, but they amplify compression artifacts greatly; (g) avoids compression artifacts amplification but results in image details blur. 

\subsubsection{Compare with \textbf{Enhancement+Deblocking}}
We further compare our approach with the manner processing outputs of enhancement algorithms with existing deblocking methods \textbf{(enhancement+deblocking)}.
After the joint processing of the enhancement algorithm and the deblocking algorithm, which is present in Fig. \ref{fig:enhance_first}, the results obtained by the methods (b), (c), (d), (e), (f), (g) cause texture detail distortion. In contrast, the proposed method outperforms the existing methods in brightness enhancement, detail texture restoration and block effect suppression, and achieves better results. 

\subsubsection{Compare with \textbf{Deblocking+Enhancement}}
We further compare our approach with the manner processing outputs of deblocking methods with existing enhancement algorithms \textbf{(deblocking+enhancement)}. 
As shown in Fig.\ref{fig:deblock_first}, results of \textbf{deblocking+enhancement} still have blocking artifacts (e, f, g). In addition, such a manner even results in color distortion (b, c, d, f), which is caused by the fact that blocking artifacts in dark images are hard for deblocking models to perceive.

\subsection{Quantitative Evaluation}
In order to verify the effectiveness of our approach, we compare our approach with existing dark image enhancement methods with different quality factors (QF), \textit{i.e.,} 90\%, 80\%, 70\%, 60\%. Table \ref{tab:compara-90-80} show the comparison results with different methods under QF 90\% and 80\%. The comparison results under QF 70\% and 60\% are listed in the appendix.

\subsubsection{Compare with \textbf{Enhancement Methods}}
We compare our approach with one conventional methods, \textit{i.e,} LIME \cite{guo2016lime}, and seven deep learning based methods, \textit{i.e.,} RetinexNet \cite{Chen2018Retinex}, RetinextDIP \cite{zhao2021retinexdip}, Kind \cite{zhang2019kindling}, EnlightenGAN \cite{jiang2021enlightengan}, ZeroDCE \cite{guo2020zero}, LLFlow \cite{wang2021low} and RUAS \cite{liu2021retinex}. It can be seen that our approach is superior to the enhancement methods under different evaluation metrics.

\subsubsection{Compare with \textbf{Enhancement+Deblocking}}
 We further compare our approach with \textbf{enhancement+deblocking} methods. For LOL and LSRW datasets, our approach outperforms \textbf{enhancement+deblocking} methods under both full-referenced and no-referenced metrics. For SICE datasets, our approach outperforms \textbf{enhancement+deblocking} methods under PSNR and PSNR-B. For SSIM, our approach also achieves comparable performance.

\subsubsection{Compare with \textbf{Deblocking+Enhancement}}
In addition, we compare our approach with \textbf{deblocking+enhancement} methods. For LOL dataset, when QF is set as 90\% and 80\%, our approach outperforms \textbf{deblocking+enhancement} methods. For LSRW dataset, our approach outperforms \textbf{deblocking+enhancement} methods. For SICE dataset, our approach still achieves comparable performance.

\subsection{Ablation Studies}
To verify the effectiveness of our proposed Multiple Latent Space Mapping method, we perform ablation studies from two perspectives. Firstly, we remove the latent mapping network and only train an encoder-decoder network to learn the enhancement. Secondly, we compare the performance of our latent mapping method with different levels, \textit{i.e.,} single-level, two-level and three-level.

\subsubsection{Ablation for Latent Mapping Network}
We remove the latent mapping network $\mathcal{M}$ to perform this ablation study. Specifically, we train an enhancement model with the same network structure as VAE, to learn the image restoration from compressed dark images to normal-light images directly. In this ablation study, we use LOL, LSRW and SICE datasets with QF 90\%, and adopt PSNR, SSIM, PSNR-B as quantitative evaluation metrics. Table \ref{tab:ablation_M} displays the quantitative results for ablation of latent mapping network. It can be seen that the latent mapping network brings a large performance improvement.

\begin{table}[ht]
\caption{\textbf{Quantitative Results for Ablation of Latent Mapping Network}}
\centering
\renewcommand\tabcolsep{15pt} 
\scalebox{0.88}{
    \begin{tabular}{cccccc}
    \toprule[1pt]
    Dataset & $\mathcal{M}$ & PSNR & SSIM & PSNR-B 
    \\ \midrule[0.5pt]
    \multirow{2}{*}{LOL} &  & 16.52 & 0.6729 & 16.57
    \\
     & $\checkmark$ & 19.16 & 0.8373 & 19.06
    \\ \midrule[0.5pt]
    \multirow{2}{*}{LSRW} &  & 15.07 & 0.4267 & 15.02
    \\
     & $\checkmark$ & 16.83 & 0.4959 & 16.81 
    \\ \midrule[0.5pt]
    \multirow{2}{*}{SICE} &  & 15.89 & 0.4626 & 15.97 
    \\
     & $\checkmark$ & 16.20 & 0.5189 & 16.89 
    \\
\bottomrule[1pt]
\end{tabular}}
\label{tab:ablation_M}
\end{table}

\subsubsection{Ablation for Multiple Latent Spaces}
In order to verify the effectiveness of our multiple latent space mapping, we perform ablation studies for the mapping levels. Specifically, we test the performance under single-level latent space mapping (only mapping on the top level), two-level latent space mapping (mapping on the top and middle levels) and three-level latent space mapping (mapping on the top, middle and bottom levels) respectively. In this ablation study, we use LOL, LSRW and SICE dataset with QF 90\%, and adopt PSNR, SSIM, PSNR-B as quantitative evaluation metrics.

\begin{table}[ht]
\caption{\textbf{Quantitative Results for Ablation of Multiple Latent Spaces}}
\centering
\renewcommand\tabcolsep{11pt} 
\scalebox{0.89}{
    \begin{tabular}{cccccc}
    \toprule[1pt]
    Dataset & Mapping level & PSNR & SSIM & PSNR-B 
    \\ \midrule[0.5pt]
    \multirow{3}{*}{LOL} & single-level & 18.65 & 0.8097 & 19.23 
    \\
     & two-level & 19.07 & 0.8146 & 19.49
     \\
     & three-level & 19.16 & 0.8373 & 19.06
    \\ \midrule[0.5pt]
    \multirow{3}{*}{LSRW} & single-level & 16.42 & 0.4856 & 16.42
    \\
     & two-level & 15.95 & 0.4830 & 16.01
     \\
     & three-level & 16.83 & 0.4959 & 16.81 
    \\ \midrule[0.5pt]
    \multirow{3}{*}{SICE} & single-level & 15.98 & 0.5104 & 16.16
    \\
     & two-level & 16.12 & 0.5090 & 16.55
     \\
     & three-level & 16.20 & 0.5189 & 16.89 
    \\
\bottomrule[1pt]
\end{tabular}}
\label{tab:ablation_k}
\end{table}

\begin{figure}[ht]
    \centering
    \includegraphics[scale=0.3]{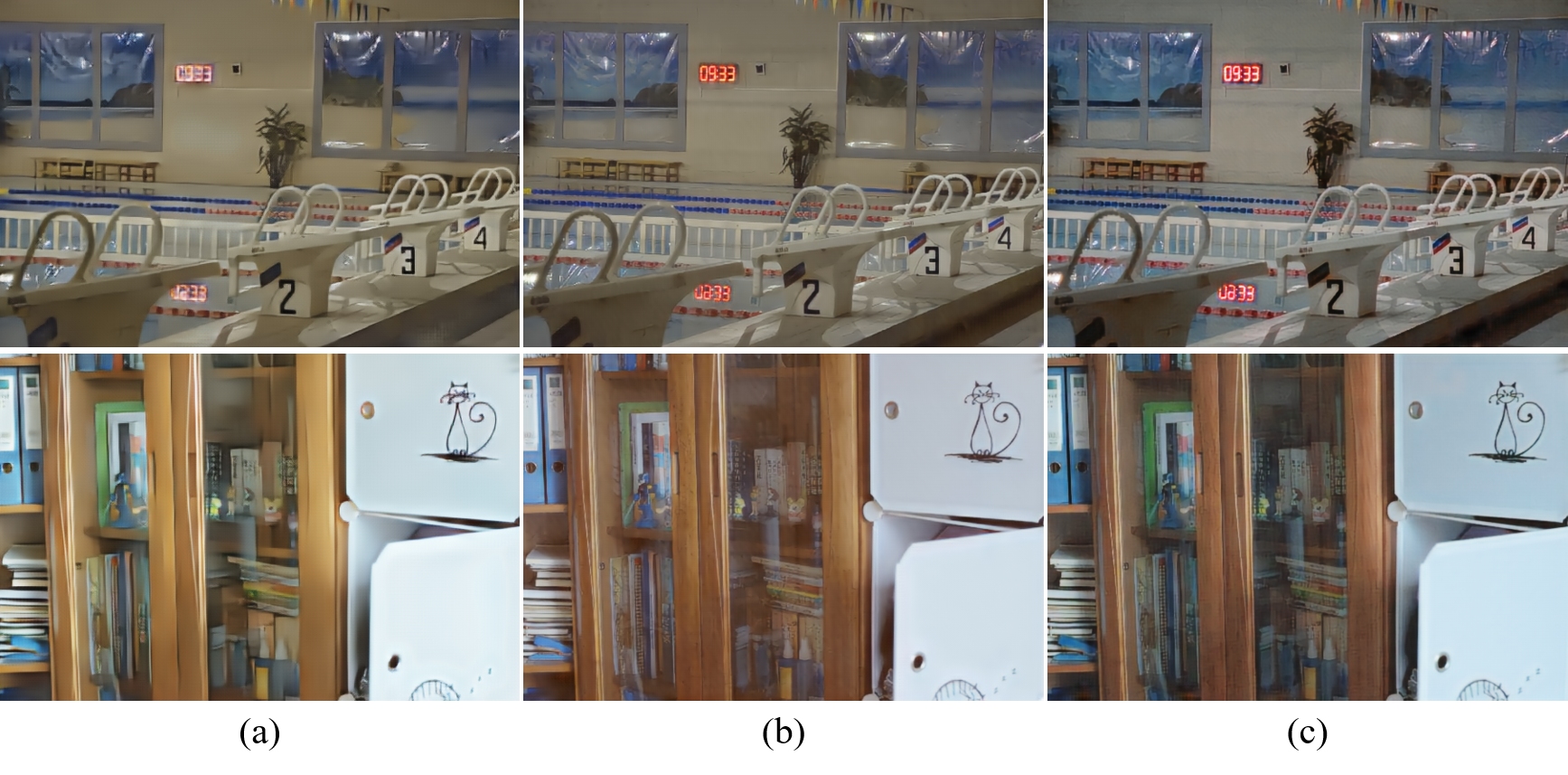}
    \caption{Ablation study of multiple mapping. (a) results of single-level mapping. (b) results of two-level mapping. (c) results of three-level mapping.}
    \label{fig:ab_k}
\end{figure}

\begin{table}[ht]\Huge
\renewcommand\tabcolsep{17pt} 
\caption{\textbf{mAP on dark face with QF 90\%.} The best values are in red. The second best values are in blue.}
\centering
\scalebox{0.32}{
    \begin{tabular}{cccc}
    \toprule[2.5pt]
    \multirow{2}{*}{Method}&\multicolumn{3}{c}{IoU threshold}
    \\
    \cmidrule(r){2-4} & 0.3 & 0.5 & 0.7
    \\ \midrule[2.5pt]
     LIME\cite{guo2016lime} & 0.188049 & 0.169873 & 0.045380
    \\
    RetinexNet\cite{Chen2018Retinex} & 0.147633 & 0.136992 & 0.039964
    \\
    KinD\cite{zhang2019kindling} & 0.159096 & 0.145566 & 0.042016
    \\
    EnlightenGAN\cite{jiang2021enlightengan} & 0.129158 & 0.120389 & 0.032258
    \\
    RetinexDIP\cite{zhao2021retinexdip} & 0.182877 & 0.165881 & 0.043670
    \\ 
    Zero-DCE\cite{guo2020zero} & 0.170194 & 0.156849 & 0.040050
    \\
    LLFlow\cite{wang2021low} & 0.184419 & 0.169579 & 0.051443
    \\
    RUAS\cite{liu2021retinex} & 0.169371 & 0.155320 & 0.042214
    \\\cmidrule(r){1-4}
    EnligntenGAN\cite{jiang2021enlightengan} + EDSR\cite{lim2017enhanced} & 0.074730 & 0.070601 & 0.018611
    \\
    ZeroDCE\cite{guo2020zero} + EDSR\cite{lim2017enhanced}  & 0.188670 & 0.172673 & 0.048338
    \\
    LLFlow\cite{wang2021low} + EDSR\cite{lim2017enhanced} & 0.160536 & 0.149057 & 0.046590 
    \\
    EnligntenGAN\cite{jiang2021enlightengan} + QGCN\cite{li-2020} & 0.039802 & 0.038792 & 0.009800
    \\    
    ZeroDCE\cite{guo2020zero} + QGCN\cite{li-2020} & 0.108781 & 0.102474 & 0.030173
     \\
    LLFlow\cite{wang2021low} + QGCN\cite{li-2020} & 0.183001 & 0.168877 & 0.050879
    \\\cmidrule(r){1-4}
    EDSR\cite{lim2017enhanced} + EnlightenGAN\cite{jiang2021enlightengan} & 0.077944 & 0.073587 &  0.020028
     \\
    EDSR\cite{lim2017enhanced} + ZeroDCE\cite{guo2020zero} & 0.082400 & 0.077467 & 0.022322
     \\
    EDSR\cite{lim2017enhanced} + LLFlow\cite{wang2021low} & 0.076276 & 0.072593 &  0.021737
     \\
    QGCN\cite{li-2020} + EnlightenGAN\cite{jiang2021enlightengan} & 0.186228 & 0.172047 & 0.048554
     \\
    QGCN\cite{li-2020} + ZeroDCE\cite{guo2020zero} & \textcolor[rgb]{0,0,1}{0.193952} & \textcolor[rgb]{0,0,1}{0.177015} & 0.047839
     \\
    QGCN\cite{li-2020} + LLFlow\cite{wang2021low} & 0.179730 & 0.167152 & \textcolor[rgb]{1,0,0}{0.052284}
     \\\cmidrule(r){1-4}
    \textbf{Ours} &  \textcolor[rgb]{1,0,0}{\textbf{0.201467}} & \textcolor[rgb]{1,0,0}{\textbf{0.185907}} & \textcolor[rgb]{0,0,1}{\textbf{0.052263}}
     \\
\bottomrule[2.5pt]
\end{tabular}}
\label{tab:dark_face}
\end{table}

The qualitative results of different levels are provided in Fig. \ref{fig:ab_k}. The single-level latent space mapping success to suppress the blocking artifacts, but results in image details blur and slight color distortion. Contrast to single-level, the two-level latent space mapping obtains better restoration results, but still causes color distortion. The three-level latent space mapping achieves great performance on details and color restoration with blocking artifacts suppression.

The quantitative results of different levels are provided in
Table \ref{tab:ablation_k}. It can be seen that three-level latent space mapping achieves the best PSNR and SSIM values, which is because the details are largely preserved by three-level latent space mapping. In LOL dataset, two-level latent space mapping achieves the best PSNR-B value, which is because two-level latent space mapping balances detail texture restoration and blocking artifacts suppression. One could choose different mapping levels according to the specific requirements on detail texture restoration and blocking artifacts suppression.

\begin{figure*}[ht]
    \centering
    \includegraphics[scale=0.42]{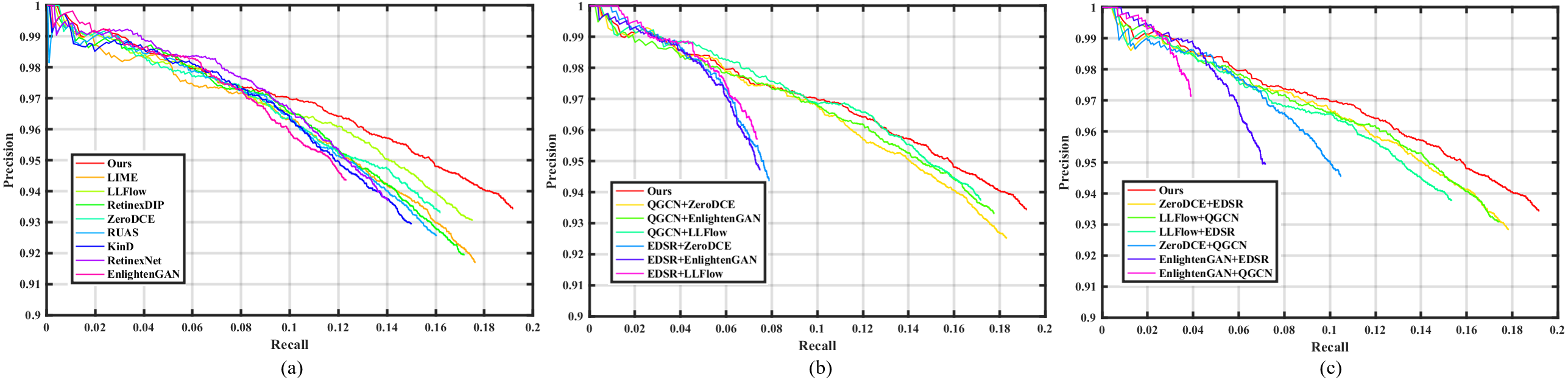}
    \caption{P-R curve comparisons on dark face detection. (a) Comparisons with \textbf{enhancement} methods; (b) Comparisons with \textbf{enhancement+deblocking} methods; (c) Comparisons with \textbf{deblocking+enhancement} methods.}
    \label{fig:pr_enhance}
\end{figure*}

\begin{figure*}[ht]
    \centering
    \includegraphics[scale=0.8]{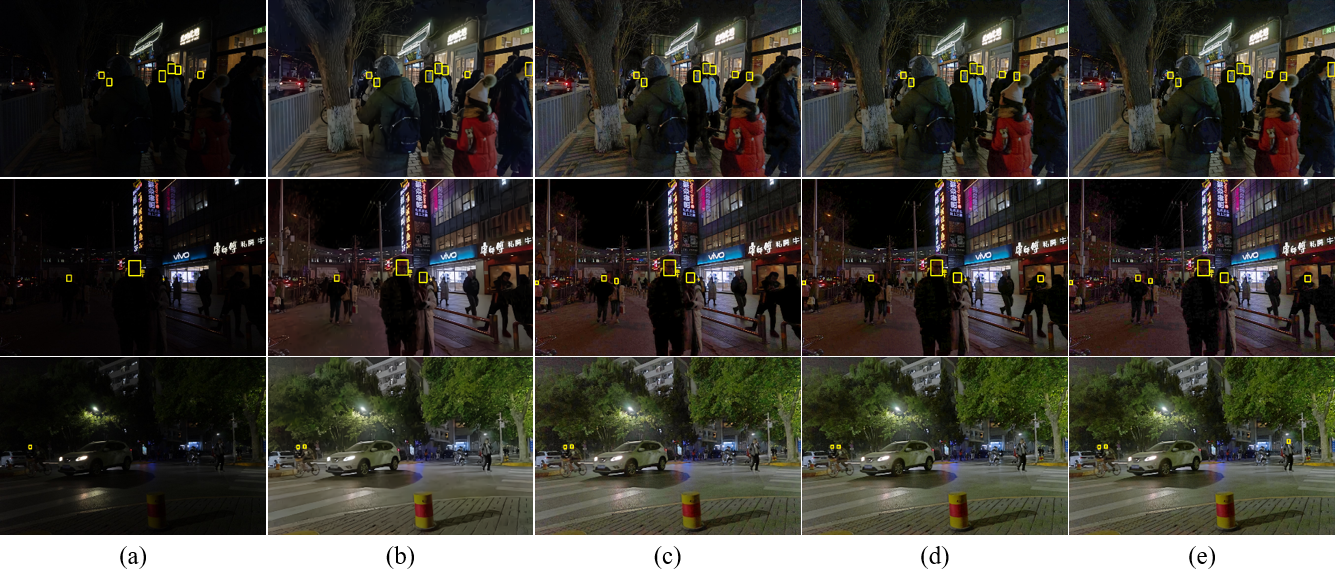}
    \caption{Visualization of dark face detection. (a) Inputs; (b) ZeroDCE; (c) ZeroDCE+EDSR; (d) QGCN+ZeroDCE; (e) Ours.}
    \label{fig:pr_enhance}
\end{figure*}



\subsection{Dark Face Detection}
To further verify the effectiveness of our compressed dark image enhancement method, we test the performance of dark image enhancement methods on the face detection task under compressed dark conditions. Specifically, we use the DARK FACE dataset that composes
10000 images taken in the dark. Since the bounding boxes of test set are not publicly available, we use the training and validation sets, which totally contain 6000 images. We use PyramidBox \cite{tang2018pyramidbox} trained on WIDER FACE dataset \cite{yang2016wider} as the face detector. For the dark face detection, we first use dark image enhancers to restore the compressed dark images, and then use PyramidBox to get the detection results.

Table \ref{tab:dark_face} shows the comparisons of mean Average Precision (mAP) under different IoU thresholds (0.3, 0.5, 0.7). Under the IoU thresholds of 0.3
and 0.5, our approach achieves the best mAP value among different methods. When the IoU threshold is set to 0.9, the mAP values of different methods get poor, and our approach still achieves comparable performance.

We further present the comparisons of precision-recall (P-R) curves under IoU threshold 0.5 in Fig. \ref{fig:pr_enhance}. It can be seen that our approach performs better in the high recall range.

 Fig.\ref{fig:pr_enhance} exhibits some visual examples of current methods and our approach. As all known, blocking artifacts are not friendly to long-distance tiny face detection. Our method performs better on long-distance tiny face detection because our method focuses on preserving details and removing blocking artifacts. Experiments of dark face detection further show the effectiveness of our approach.

\section{Conclusion}
We propose a significant task named compressed dark image enhancement and propose a novel method for this task. We have observed that compression artifacts would be enhanced by existing dark enhancement methods. Based on the observation, we propose to perform the restoration from compressed dark images to normal-light images in latent space. Specifically, we propose a novel two-branch latent mapping network based on multi-level VAE. Comprehensive experiments show the effectiveness of our approach. Furthermore, we perform experiments under different quality factors and show that our performance is comparable to existing methods.

\bibliographystyle{IEEEtran}
\bibliography{main}



\end{document}